\documentclass[a4paper,fleqn]{cas-dc} 
\usepackage[numbers]{natbib}

\usepackage{amssymb}
\usepackage{amsmath}
\usepackage{amsthm}
\usepackage{etoolbox}
\usepackage{amsfonts}
\usepackage{mathtools}
\usepackage{verbatim}
\usepackage{booktabs} 
\usepackage{siunitx}  
\usepackage{listings}
\usepackage{amsthm}
\newtheorem{definition}{Definition}[section]
\newtheorem{theorem}{Theorem}[section]
\usepackage{algorithm}
\usepackage{algorithmic}
\usepackage{fancyvrb}
\usepackage{graphicx}
\usepackage{subcaption}
\usepackage{dashbox}
\usepackage{hyperref}
\usepackage{newfloat}
\usepackage{caption}

\usepackage{multirow}

\usepackage{pifont}
\def\tsc#1{\csdef{#1}{\textsc{\lowercase{#1}}\xspace}}
\tsc{WGM}
\tsc{QE}
\tsc{EP}
\tsc{PMS}
\tsc{BEC}
\tsc{DE}

\begin{document}
\let\WriteBookmarks\relax
\def\floatpagepagefraction{1}
\def\textpagefraction{.001}
\shorttitle{Fast and Accurate Anomaly Detection in Time Series}

\title [mode = title]{Fast and Accurate Anomaly Detection in Time Series}




\author[1]{Emanuele Mele}[orcid=0000-0001-9297-1434]
\shortauthors{Emanuele Mele et al.}



\ead{emanuele.mele@unisalento.it}


\credit{Conceptualization, Methodology, Software, Validation, Formal analysis, Investigation, Writing - Original Draft, Writing - Review \& Editing, Visualization}

\affiliation[1]{organization={University of Salento},
            addressline={Via per Monteroni}, 
            city={Lecce},
            postcode={73100}, 
            state={},
            country={Italy}}

\author[1]{Massimo Cafaro}[orcid=0000-0003-1118-7109]
\cormark[1]


\ead{massimo.cafaro@unisalento.it}


\credit{Conceptualization, Methodology, Validation, Formal analysis, Investigation, Writing - Original Draft, Writing - Review \& Editing}

\cortext[1]{Corresponding author}


\author[1]{Angelo Coluccia}[orcid=0000-0001-7118-9734]


\ead{angelo.coluccia@unisalento.it}


\credit{Conceptualization, Methodology, Validation, Formal analysis, Investigation, Writing - Review \& Editing}

\author[1]{Italo Epicoco}[orcid=0000-0002-6408-1335]


\ead{italo.epicoco@unisalento.it}


\credit{Methodology, Validation, Formal analysis, Investigation, Writing - Review \& Editing}

\begin{abstract}
Anomaly detection is a critical and evolving field in Machine Learning, with applications targeting different domains such as cybersecurity, finance, healthcare, manufacturing and IoT (Internet of Things) systems. Traditionally, anomaly detection algorithms have been designed using both supervised and unsupervised learning paradigms. The fundamental challenge in real-world anomaly detection scenarios is related to the inherent class imbalance (anomalies are typically rare) and, for supervised methods, to the scarcity of labelled anomalous data. Indeed, labelling is both expensive and time-consuming. Conversely unsupervised methods do not require labelling, but may suffer from high false positive rates when deployed in safety-critical applications. In this work we introduce a novel unsupervised algorithm for anomaly detection in time series based on the Haar discrete wavelet and a suitably designed $t$-test. We establish the theoretical foundation of the proposed $t$-test and, through extensive experimentation across 343 datasets, demonstrate that our algorithm outperforms state-of-the-art unsupervised and self-supervised benchmarks.
\end{abstract}



            
\maketitle

\section{Introduction}
\label{introduction}
Anomaly detection represents a cornerstone of Machine and Deep Learning research, finding extensive utility across a diverse array of industrial and scientific domains. In cybersecurity, it can be employed to detect unauthorised breaches, while in finance it proves valuable for identifying fraud in credit card transactions. The healthcare sector benefits from its ability to flag the onset of rare medical conditions, and manufacturing leverages it to spot microscopic production defects. Beyond these, anomaly detection plays a role in IoT systems for detecting hardware failures of monitored sensors and extends even to sociology, where it can identify cyberbullying and vulnerable users exhibiting signs of depression or suicidal tendencies in social networks, and to astronomy, where it aids in the discovery of very rare galaxy types or exoplanets. Anomalies are not necessarily associated with negative abnormal behaviour, in which they act as a proxy for a problem to be solved: Some data observations may actually be a sign of novel data or related to a positive process outcome.  

 Anomalies, also known as outliers, discordants or deviants, are informally defined as data that differ widely from the other data (which are considered to be regular) in a sample, dataset or time series \cite{Grubbs01021969}. In particular, an anomaly has been defined as "an observation which deviates so much from other observations as to arouse suspicions that it was generated by a different mechanism" \cite{Hawkins1980, bsp}.
 
 Timely detection of anomalies is fundamental in many applications, in which we do not expect the occurrence of such abnormal observations. The assumption is that all the observations should be regular, i.e., exhibit normal behaviour with regard to a predefined model, prior belief or hypothesis. Normal data are also called inliers in the literature. We expect data exhibit the stationarity property or, alternatively, that changes in the observations are caused by regular cyclic patterns (such as seasonal trends, etc).
 
 Hence, algorithms for anomaly detection typically begin by trying to establish what can be considered as normal behaviour, typically assuming a distribution or estimating one through historical data, taking into account the relevant features of the data. However, this task may be challenging, owing to several factors, e.g. any attempt to subsume any possible normal behaviour may be overly restrictive. Moreover, there may be many observations near the limiting border separating inliers from outliers, making it difficult to establish precisely which observations are normal and which ones are anomalies, thus the border itself may easily be imprecise. Another difficulty lies in the possible evolution, over time, of what constitutes a normal behaviour.
 
 A common approach consists in treating anomaly detection as a supervised binary classification task, in which data observations are classified as either inliers or outliers. However, owing to the underlying class distribution and to a dramatic class imbalance, classification algorithms are usually bound to failure. Since they are characterised by a huge number of false negatives, achieving high accuracy is challenging. In general, a supervised model is only as good as the diversity of its training labels; in the context of anomaly detection, supervised models often fail to identify novel (a.k.a. zero-day) anomalies that were not present in the training set. Moreover, a supervised model facing the class imbalance problem may easily achieve high accuracy by simply ignoring the minority class, a risk that must be taken into account and mitigated as much as possible, e.g., by using techniques such as SMOTE (Synthetic Minority Oversampling Technique) \cite{Chawla2002} or ADASYN (Adaptive Synthetic Sampling) \cite{Haibo2008}.
 
 A different class of algorithms does not output a binary label but, rather, a numerical outlier score that quantifies the degree of outlierness of each observation. The outlier score of an observation must be compared to a suitable threshold in order to determine if the observation is an outlier or not (this applies to classifiers as well, which determine the output binary label by a thresholding mechanism applied to their specific outlier score). 
 
It is worth recalling here that in the literature the terms outlier and anomaly are mostly interchangeable, but notable exceptions do exist \cite{Aggarwal2001}, with some authors referring to outliers as noise (uninteresting deviations from inliers) in order to distinguish them from true anomalies which are, instead, considered as interesting deviations from inliers, i.e., significant events that deserve to be analysed.
 
 In this work, we focus in particular on anomaly detection in univariate time series, a sequence of data points ordered along the time. 
 We adopt an unsupervised approach, since this kind of algorithms has been shown to be the most flexible and effective in the literature, and is therefore widely deployed. The key assumption is that anomalies are not only rare, they also differ significantly from the rest of the data. Since unsupervised models do not rely on explicitly provided labels, they discover and exploit the intrinsic structural properties of the data (e.g. the density, distance, or isolation) to detect anomalies, and are quite good at discovering novel anomalies. On the other hand, being unsupervised these models may be susceptible to high false positive rates. The lack of supervision may lead to flagging as anomalies those data that, despite being unusual, are still valid, legitimate data.
 
Our proposed approach, DWTt-test, leverages the multi-resolution analysis of the Haar Discrete Wavelet Transform (DWT) to decompose univariate time series into coarse and detail components. This decomposition allows the algorithm to simultaneously monitor underlying trends and localised high-frequency variations. By applying a sliding window mechanism across these levels, we employ a rigorously derived, ad-hoc $t$-test to assign an anomaly score to each observation based on its statistical significance. Unlike many deep learning-based approaches, our method is entirely unsupervised and does not require a training phase, making it robust to the inherent class imbalance and label scarcity typical of real-world scenarios.

The primary contributions of this work are summarised as follows:

\begin{itemize}
    \item We introduce DWTt-test, an unsupervised anomaly detection algorithm that combines DWT multi-level decomposition with a tailored statistical framework;
    \item We provide a formal mathematical derivation proving that our ad-hoc $t$-score follows a Student's $t$-distribution under the operational settings of the algorithm;
    \item We conduct a large-scale evaluation across 343 diverse datasets, demonstrating that our approach outperforms state-of-the-art unsupervised and self-supervised models;
    \item We prove that the algorithm exhibits a strict linear time complexity, $\mathcal{O}(N)$, ensuring high throughput and suitability for real-time deployment on resource-constrained hardware.
\end{itemize}

The rest of this paper is organised as follows. Section \ref{related} discusses related work in the field of anomaly detection. Section \ref{algorithm} provides a detailed presentation of the DWTt-test algorithm and its theoretical derivation. Section \ref{results} reports the results of our extensive experimental evaluation. Finally, Section \ref{conclusions} draws the conclusions and outlines future research directions.
 
 \section{Related Work}
 \label{related}
Anomaly detection has been extensively studied and many different algorithms have ben designed for this task, initially based on statistics and later on advanced Machine and Deep Learning representations. Here, we recall the most important related work without demanding to be exhaustive, owing to space. The interested readers can find additional details in several reviews such as \cite{Kesharwani2024, Liso2024, Li2025, SHI2026}.

 We shall discuss several algorithms, representatives of different approaches.

 \subsection{Distance-based Algorithms} 
 This category encompasses various algorithms that identify outliers through distance-based metrics, including Euclidean, Minkowski, and Mahalanobis distances. While the Euclidean distance treats all variables as independent and equally weighted, the Mahalanobis distance—frequently employed in multivariate scenarios—accounts for the covariance between variables. Specifically, it scales the feature space to ensure that variables characterised by high variance or significant correlation do not disproportionately influence the distance calculation. In this context, the Jaccard and the cosine similarity have been used to devise algorithms. Using the sum of distances from all the observations, outliers are those points farthest from the observations in the dataset. Alternatively, outliers are those observations whose nearest neighbour is farthest.  A variant is based on considering the set $\mathcal{N}$ of $k$ nearest neighbours ($k$-NN algorithm) and the average distance (e.g., either the median or the sum of the distances) to the observations belonging to $\mathcal{N}$ \cite{knnrd}.

\subsection{Unsupervised Algorithms}
The Local Outlier Factor (LOF) algorithm \cite{Breunig2000} was designed to address a vulnerability of distance-based algorithms, such as $k$-NN, when applied to datasets characterised by the presence of clusters of different density. A significant limitation of global distance-based metrics arises when an observation's absolute distance to its neighbours is smaller than the inter-point distances within a nearby sparse cluster. Despite this proximity, the point may still be characterised as an outlier if it resides adjacent to a high-density cluster, where its relative local density is significantly lower than that of its surroundings. This is addressed by LOF computing a point's local density in relation to its neighbours' local densities. Then, a point is flagged as a local outlier if its density is much lower than that of its neighbours (i.e., its LOF score is much greater than one). 

The main idea behind clustering-based algorithms is that once the dataset has been clustered, all the observations lying outside the detected clusters are potential outliers. Some algorithms treat as anomalies those observations near or on the boundary of a cluster. Depending on the clustering algorithm used, outliers can also be identified directly. As an example, the DBSCAN (Density-Based Spatial Clustering of Applications with Noise) clustering algorithm \cite{Ester1996} directly determines the outliers (called noise points). CURE (Clustering Using Representatives) \cite{Guha1998} represents each cluster using a set of so-called representative points carefully selected (from those well scattered near the cluster, which are then moved towards the center of the cluster). Two clusters are merged if the minimum distance among any pair of the corresponding representatives is below a predefined threshold, and outliers are detected considering how fast a cluster is growing, since clusters made of inliers should grow relatively fast: Therefore, a slowly growing cluster is with high probability made of potential outliers. Another algorithm, proposed in \cite{BFR1998}, also identifies outliers explicitly. The data, assumed to be large enough to not fit in the available memory, is handled one chunk at a time, and the algorithms maintains three sets of points. These are the discard set, the compressed set and the retained set. The discard set is the set of points that actually belong to a cluster: These are summarised through cleverly designed statistics and then discarded. The compressed set includes those points that are close to one another, but not close enough to one of the clusters (otherwise they would have been assigned to the discard set). Finally, the retained set is the set of remaining points. These cannot be assigned to any cluster, nor are they close enough to one another: These are isolated points, i.e., the set of outliers. In the last iteration of the algorithm, one can decide what to do with both the compressed and retained sets. They may be assigned to the available clusters, or treated as outliers. Alternatively, one can cluster only the compressed set and provide the retained set as one of the outputs. 

Isolation Forest \cite{Liu2008} differs from traditional anomaly detection algorithms in a fundamental way: Instead of first characterising inliers and identifying outliers as observations deviating from that characterisation, it strives to isolate the anomalies while discovering the dataset underlying structure. The iForest algorithm creates a forest of trees called iTrees, recursively choosing a random feature and partitioning the data using a random value belonging to the range of the selected feature. Anomalies are easier to isolate this way and are characterised by a smaller path length from the root node, and the anomaly score is indeed based on the average path length. The algorithm has been later improved by removing a bias related to the anomaly score, which stems from the fact that decision boundaries are horizontal or vertical. The resulting algorithm \cite{Hariri2021} is called Extended Isolation Forest, and leverages decision boundaries characterised by a random slope.

One-Class Support Vector Machine (OCSVM) \cite{Schölkopf2001} is a variant of the Support Vector Machine (SVM) algorithm, designed for outlier detection. Like the SVM, OCSVM learns a decision boundary (i.e., a separating hyperplane) for a dataset; typically, this is a nonlinear boundary based on a Radial Basis Function kernel. The OCSVM algorithm works by determining a separating hyperplane that ideally puts all the dataset observations on one side of it, taking into account the maximum allowed training error. Next, subsequent observations shall be classified as anomalies if they fall outside the learned decision boundary, and as inliers otherwise.

Autoencoders \cite{Aggarwal2023} are neural networks that consist of an encoder and a decoder network. While the encoder provides a low-dimensional representation of the input data through a latent space, achieved using hidden layers of reduced dimensionality with regard to the input space, the decoder performs the opposite task, expanding the low-dimensional data from the latent space to the original dimension of the input space. This network aims at reconstructing the input, but this task is complicated by the fact that the hidden layers are actually operating a lossy compression, retaining the most useful information and discarding the rest. The encoded information is then transformed back into the input from the decoder network. When the number of features is sufficiently high, this form of compression allows efficiently learning the dataset: To the extent possible, an input affected by noise will be correctly reconstructed. Therefore, denoising is a typical application of autoencoders. Moreover, they can  be used for anomaly detection as well: When encoded and decoded, outliers typically cannot be correctly reconstructed, since they lose substantial information. Variants include variational and adversarial autoencoders \cite{Somepalli2021}.

Deep Support Vector Data Description (DeepSVDD) \cite{Ruff2018} combines the feature extraction power of deep neural networks (it extracts a lower-dimensional data representation) with the boundary-learning capabilities of OC-SVM, an approach known as Deep One-Class Classification. The key idea is to minimise, through SVDD, the volume of the enclosing hypersphere on the extracted lower-dimensional representation. It integrates two popular anomaly detection techniques: Training a single neural network to simultaneously extract a lower-dimensional data representation and a support vector data descriptor (SVDD) that minimises the volume of the enclosing hypersphere on this lower-dimensional representation. In practice, normal data instances are concentrated in this hypersphere of minimum volume. DeepSVDD has been shown to perform better than the baselines, and several variants have been proposed. However, it is subject to a critical issue known as "hypersphere collapse" during training. When the neural network approaches the trivial solution of all-zero weights, hypersphere collapse takes place. To contrast this, the authors of DeepSVDD limited the algorithm's performance and efficacy by imposing a number of restrictions on the deep network's architecture, such as fixing the hypersphere center and setting network biases to zero. Deep Autoencoding SVDD \cite{Hojjati2024} and Contrastive Deep SVDD \cite{Xing2023} are recent variants solving the hypersphere collapse problem.

\subsection{Supervised Algorithms}
Shallow neural networks have been explored in an attempt to devise anomaly detection algorithms, but they are severely limited by their intrinsic inability to model sequential/temporal dependencies and relationships. Deep neural networks, such as Recurrent Neural Networks (RNN) and Long Short Term Memory (LSTM), can instead model sequential and temporal patterns, and have been used for anomaly detection \cite{Su2019, Tran2019, Lobach2024}. Generative Adversarial Networks (GANs) can be trained to detect anomalies on the basis of normal data generation, but their application appears to be limited to some domains, owing to exhibited instability problems \cite{SCHLEGL2019, Li2019}.

Recently, transformers \cite{Vaswani2017} have shown their effectiveness in handling complex temporal dependencies using an attention mechanism (providing the strength of the connections among the input tokens) that allows encoding and processing the input using a set of parameters (related to so-called values, queries and keys) that does not increase with the input length. Moreover, transformers greatly improve several aspects with regard to RNN: Besides the ability to handle long-range dependencies, they do not suffer the gradient vanishing and exploding problems, require fewer training steps and the training can be parallelised; however, inference is slow since it scales quadratically with the input sequence length. Transformers have been used autoregressively in time series forecasting problems, better exploiting long-term dependencies \cite{Benali2024} and have been used in the context of anomaly detection \cite{Xu2022, Xu2022anomaly, Baidya2023, Li2025, Shrestha2025, dcdetector}. However, owing to the high quadratic cost incurred by the underlying attention mechanism, State-Space Models (SSM) are becoming popular. 

SSM can be used to process sequences, and model a system through its possible states. Mathematically, SSM are characterised by a state and an output equation modelling a dynamic system of interest. These models can use a convolutional representation in the training phase, allowing parallelisation, and can use a recurrent representation during inference, leading to efficient inference (we recall here that Transformers are not efficient at inference time). Among these models, MAMBA is now becoming ubiquitous \cite{Lahoti2026}. In particular, MAMBA allows selectively filtering non relevant information: The state size is much smaller with regard to a transformer, and is therefore more efficient. Moreover, the design also includes a hardware-aware algorithm for efficient storage of intermediate results. While the training and inference complexity of a transformer are respectively $O(n^2)$ and $O(n)$ where $n$ is the length of the input sequence, the correspondning complexities for MAMBA are $O(n)$ and $O(1)$. Recently, MAMBA has been used for anomaly detection \cite{SELLAM2025}. The authors used the classic MAMBA model in conjunction with the association discrepancy also used in \cite{Xu2022anomaly, Li2025, Shrestha2025}. This is defined in \cite{Xu2022anomaly} as the symmetric Kullback–Leibler divergence which measures the distance between the series-association and the prior-association distributions. In practice, the series-association is the association distribution of each time point as discovered by a transformer self-attention map. Normal data points may find associations of this kind with almost all the data points in the time series. In contrast, the prior-association distribution is related to the idea that anomalies, being rare, cannot exhibit strong associations with the whole time series. Therefore, anomalies associations are typically confined to near, adjacent data points. In \cite{Xu2022anomaly} the authors proposed a minimax strategy for association learning, to overcome some problems that may originate from direct maximisation of association discrepancy. 
 
\section{Proposed Algorithm}
\label{algorithm}

\begin{algorithm}
	\footnotesize
    \caption{DWTt-test}
    \label{alg:dwtt-test}
    \begin{algorithmic}[1]
        \REQUIRE Normalised time series \(\mathbf{x}\), time series size $N$, window-size $w$, max level $\hat{L}$, detection threshold $\epsilon$
        \ENSURE Outlier scores $\mathbf{z}_{0}$

        \STATE Increase the dimension of \(\mathbf{x}\) from \(N\) to \(M=2^{\lceil \text{log}_2(N) \rceil}\)

        \FOR{\( l = \hat{L}\) \TO \(1\)}
            \STATE \( \mathbf{d}_l,\mathbf{c}_l = \) HaarWavelet\((\mathbf{x},l)\) \COMMENT{Compute the DWT across \(l\) levels}
            \STATE \(w_l = w(\hat{L}-l+1)\)
            \STATE Build \(\mathbf{D}_l, \mathbf{C}_l \in \mathbb{R}^{(\frac{M}{2^l}-w_l+1) \times w_l}\) by sliding a window of size \(w_l\) over \(\mathbf{d}_l,\mathbf{c}_l\)
            \STATE Compute \(\mathbf{d}^{\text{avg}}_l,\mathbf{c}^{\text{avg}}_l \in \mathbb{R}^{\frac{M}{2^l}-w_l+1}\) by averaging the rows of \(\mathbf{D}_l, \mathbf{C}_l\)
            \STATE Compute the standard deviations \(S_l^d\) and \(S_l^c\) from $\mathbf{d}^{\text{avg}}_l,\mathbf{c}^{\text{avg}}_l$
            \STATE Initialise the outlier vectors \(\mathbf{o}^d_l, \mathbf{o}^c_l \in \mathbb{R}^{\frac{M}{2^l}-w_l+1}\) to zero
            \FOR{$i=0$ to $\frac{M}{2^l}-w_l$}
                \STATE $t^d_{i,l,q-1} = \frac{d^\text{avg}_{i,l}}{S_l^d}$ \COMMENT{$t$-score for the detail coefficients}
                \STATE $t^c_{i,l,q_l-1} = \frac{c^\text{avg}_{i,l}}{S_l^c}$ \COMMENT{$t$-score for the coarse coefficients}
                \IF{$\mathbb{P}(t^d > |t^d_{i,l,q_l-1}|) \cdot 2 > \epsilon$}
                    \STATE $o^d_{i,l} = 1$
                \ENDIF
                \IF{$\mathbb{P}(t^c > |t^c_{i,l,q-1}|) \cdot 2 > \epsilon$}
                    \STATE $o^c_{i,l} = 1$
                \ENDIF
            \ENDFOR
            \STATE \(\mathbf{o}_l=\mathbf{o}^d_l+\mathbf{o}^c_l\)
            \IF{$l = \hat{L}$}
                \STATE Initialise $\mathbf{z}_{l+1} \in \mathbb{R}^{\frac{M}{2^{l+1}}}$ to NULL
            \ELSE
                \STATE $\mathbf{z}_{l+1} = \mathbf{z}_l$
            \ENDIF
            \STATE $\mathbf{z}_l = \text{BuildTree(}M,l,w_l,\hat{L}, \mathbf{o}_l,\mathbf{z}_{l+1}$) 
        \ENDFOR
    
    \STATE Build \(\mathbf{D}_0\in \mathbb{R}^{(M-w+1) \times w}\) by sliding a window of size \(w\) over \(\mathbf{x}\)
    \STATE Compute \(\mathbf{d}^{\text{avg}}_0 \in \mathbb{R}^{M-w+1}\) by averaging the rows of \(\mathbf{D}_0\)
    \STATE Compute the standard deviations \(S^d\) from $\mathbf{d}^{\text{avg}}_0$
    \STATE Initialise the outlier vectors \(\mathbf{o}^d_0, \in \mathbb{R}^{M-w+1}\) to zero
    \FOR{$i=0$ to $M-w$}
    \STATE $t^d_{i,0,q-1} = \frac{d^\text{avg}_{i,0}}{S_0^d}$ \COMMENT{$t$-score for the detail coefficients}
    \IF{$\mathbb{P}(t^d > |t^d_{i,0,q-1}|) \cdot 2 > \epsilon$}
        \STATE $o^d_{i,0} = 1$
    \ENDIF
    \ENDFOR
    \STATE \(\mathbf{o}_0=\mathbf{o}^d_0\)
    \STATE $\mathbf{z}_{1} = \mathbf{z}_l$
    \STATE $\mathbf{z}_0 = \text{BuildTree(}M,1,w,\hat{L}, \mathbf{o}_0,\mathbf{z}_{1}$) 
    \STATE Remove from \(\mathbf{z}_0\in \mathbb{R}^M\) the additional elements added at the beginning of the algorithm to restore the original dimension \(N\)

    \RETURN \(\mathbf{z}_0\)
    \end{algorithmic}
\end{algorithm}

Our unsupervised DWTt-test algorithm (Alg. \ref{alg:dwtt-test}) exploits the Discrete Wavelet Transform (DWT) and an ad-hoc $t$-test for anomaly detection in univariate time series.

\subsection{Wavelet Transform}
\begin{figure*}
    \centering
\includegraphics[width=0.80\textwidth]{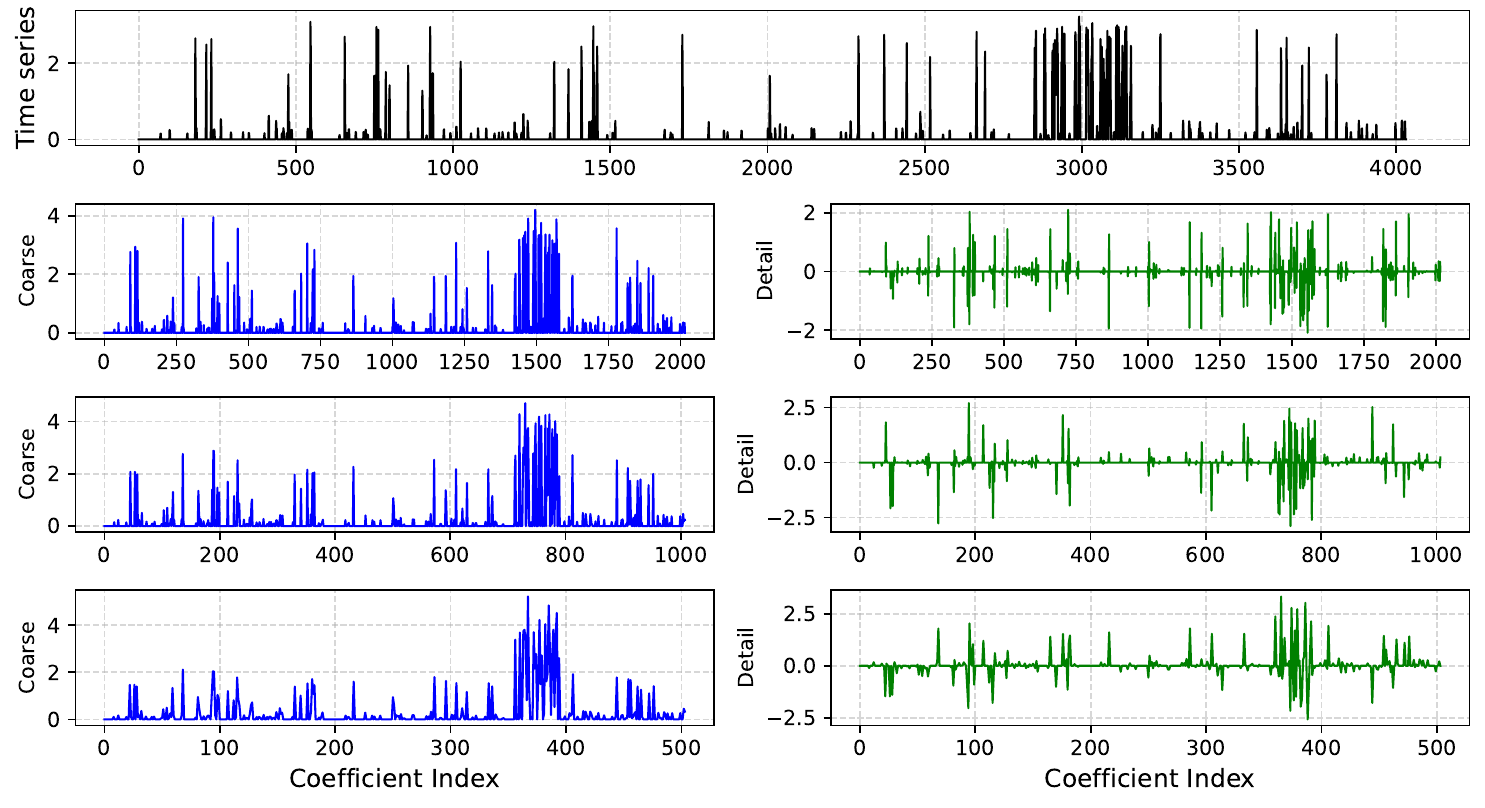}
    \caption{Multi-level Haar Wavelet decomposition ($\hat{L}=3$) of a representative signal from the NAB dataset collection. The top panel illustrates the original univariate time series. The subsequent rows display the iterative decomposition process: The left column shows the approximation or coarse coefficients, which capture the low-frequency trend, while the right column isolates the high-frequency detail coefficients, which are highly responsive to abrupt changes and anomalous behaviours. Note the progressive downsampling of the coefficient index at each decomposition level.}
    \label{fig:wavelet}
\end{figure*}
Wavelet transforms \cite{meyer1992wavelets} enable the representation of time series signals through the corresponding extracted information. Unlike (short-time) Fourier transforms, which are hampered by a constant trade-off between frequency and time resolution, wavelet transforms excel at extracting precise, time-localised frequency data. This process involves two primary operations on a "mother wavelet": Scaling (compressing or expanding the wave) and shifting (moving it along the time axis). The transform is then computed by taking the scalar product of these modified wavelets and the original signal. The Discrete Wavelet Transform (DWT) is the result of sampling the wavelets at discrete intervals. Crucially, the DWT decomposes the signal into two distinct components: The \textit{coarse} (or approximation) component, which captures the underlying low-frequency trend, and the \textit{detail} component, which isolates high-frequency, localised variations (see Figure \ref{fig:wavelet}). The DWT is highly favoured in practical applications due to its linear computational efficiency. Specifically, the current version of DWTt-test utilises a decimating DWT based on Haar wavelets for signal analysis.

Since our algorithm is based on DWT, time series must have a length equal to a power of two to work properly. Imposing this constraint on the time series is not an issue, owing to the fact that it is always possible to artificially increase the length of the time series (a padding operation) for instance by copying the last segment of the original time series. Obviously, when the model is run, only the anomaly scores of the original time series are considered. Hence, given a time series \(\mathbf{x} \in \mathbb{R}^N\), a pre-processing step is required to increase the size of the signal to length \( M=2^{\lceil \text{log}_2(N) \rceil} \). The Haar wavelet transform can compute up to \(L = \text{log}_2(M)\) levels of detail and coarse coefficients. We denote the detail coefficients as \(d_{k,l}\) and the coarse coefficients as \(c_{k,l}\) where \(k\) is the time index and \(l\) is the wavelet's level. Specifically, level \(l = 0\) has only detail coefficients and this level is equivalent to the original time series of length \(M\), while the level \(L\) has detail and coarse coefficients of length one. In general, the size of the time series at level \(l\) is equal to \( \frac{M}{2^l} \). DWTt-test uses both the detail and coarse coefficients in addition to the original signal at level \(l=0\). 
Even though the number of possible levels is \(L\), the use of all these levels is not strictly required. Hence, for our algorithm \(\hat{L}\) is a hyper-parameter and is usually chosen as \(1 \leq \hat{L} \leq L\). All the subsequent operations are performed starting from the highest level \(l=\hat{L}\) down to the level \(l=0\).

\subsection{Sliding Windows}
The proposed algorithm aims to detect outliers across the DWT levels. Even though the DWT is able to compress temporal information across levels, it is important to analyse these levels locally by using sliding windows to catch the local behaviour of the time series. For each DWT level \( l \in {1,2,...,\hat{L}}\), given a window of length \(w\) and a stride equal to one (maximum overlapping), the algorithm rearranges the windows into two matrices \(\mathbf{D}
_{l}, \mathbf{C}_{l} \in \mathbb{R}^{q_l \times w_l}\)  for detail and coarse coefficients respectively, where \(q_l = \frac{M}{2^l} - w_l +1\) is the number of windows. Next, the average of each row of the matrices is computed, leading to two new vectors \(\mathbf{d}^\text{avg}_{l},\mathbf{c}^\text{avg}_l\in \mathbb{R}^{q_l}\). In the final part of the algorithm, level \(l=0\) is also used. Taking into account that level \(l=0\) has only detail coefficients, all the previous operations are performed on the detail matrix only. Since the size of the time series varies across levels, the algorithm adapts the dimension of the window during iterations. In particular, the window size adjustment is performed as \(w_l = w(\hat{L}-l+1)\) where \(l \in {\hat{L},\hat{L}-1,..., 0}\), hence setting \(w_{\hat{L}} = w\) and \(w_{0} = w (\hat{L}+1)\). 

\subsection{$t$-Test Detection}
After the computation of the \(\mathbf{d}^\text{avg}_{l},\mathbf{c}^\text{avg}_l\) vectors and by choosing an appropriate value of \(w\), each element  \({d}^\text{avg}_{i,l}, {c}^\text{avg}_{i,l}\) with \(i\in {1,2,...,q_l}\) can be approximately considered as a sample from a Gaussian random variable thanks to the central limit theorem.
Starting from this assumption, the idea is to compute the \textit{t}-score of each element \({d}^\text{avg}_{i,l}, {c}^\text{avg}_{i,l}\) using the following formulas: 

\begin{align}
    t^d_{i,l,q_l-1} &= \frac{d^\text{avg}_{i,l} - \mu^d}{S_l^d}, & 
    t^c_{i,l,q_l-1} &= \frac{c^\text{avg}_{i,l} - \mu^c}{S_l^c},
    \label{eq:tscores}
\end{align}

where $t^d_{i,l,q_l-1}$ and $t^c_{i,l,q_l-1}$ are the $t$-scores for the $i$-th element of the $\mathbf{d}^\text{avg}_{l}$ and $\mathbf{c}^\text{avg}_{l}$ vectors, respectively. These scores follow a Student's $t$-distribution with $q_l-1$ degrees of freedom. The parameters $\mu^d$ and $\mu^c$ represent the population means, which are considered equal to $0$ after the standardisation of the time series. The sample variances, $(S_l^d)^2$ and $(S_l^c)^2$, are computed as follows:

\begin{align*}
    (S_l^d)^2 &= \frac{1}{q_l - 1} \sum_{i=1}^{q_l} (d^\text{avg}_{i,l} - \bar{d}^\text{avg}_{l})^2, \\
    (S_l^c)^2 &= \frac{1}{q_l - 1} \sum_{i=1}^{q_l} (c^\text{avg}_{i,l} - \bar{c}^\text{avg}_{l})^2
\end{align*}

where $\bar{d}^\text{avg}_{l}$ is the average of the $\mathbf{d}^\text{avg}_{l}$ vector, and $\bar{c}^\text{avg}_{l}$ is the average of the $\mathbf{c}^\text{avg}_{l}$ vector. 
Once the \textit{t}-scores are computed, it is possible to exploit the \(\mathbf{t}^d_{l,q_l-1}, \mathbf{t}^c_{l,q_l-1}\) vectors to determine the outliers for level \(l\) by setting a significance value \(\epsilon \). In particular, the algorithm employs a two-tailed test to evaluate the statistical significance of the coefficients. Following standard probability notation let $t^d \sim t(q_l-1)$ and $t^c \sim t(q_l-1)$ represent the random variables following a Student's $t$-distribution with $q_l-1$ degrees of freedom. The algorithm considers the $i$-th element an outlier if its specific observed $t$-scores, $t^d_{i,l,q_l-1}$ and $t^c_{i,l,q_l-1}$, yield a $p$-value lower than the predefined significance thresholds, satisfying the conditions:

\[ \mathbb{P}(t^d > |t^d_{i,l,q_l-1}|) \cdot 2 < \epsilon, \]

and

\[ \mathbb{P}(t^c > |t^c_{i,l,q_l-1}|) \cdot 2 < \epsilon. \] This process generates two outlier vectors $\mathbf{o}^d_{l}, \mathbf{o}^c_{l} \in \mathbb{R}^{q_l} $, whose elements are defined as:

\begin{center}
\begin{minipage}{0.45\textwidth}
\[
o^d_{i,l} = 
\begin{cases} 
1 & \text{if } \mathbb{P}(t^d > |t^d_{i,l,q_l-1}|) \cdot 2 > \epsilon \\
0 & \text{otherwise}
\end{cases},
\]
\end{minipage}
\hfill
\begin{minipage}{0.45\textwidth}
\[
o^c_{i,l} = 
\begin{cases} 
1 & \text{if } \mathbb{P}(t^c > |t^c_{i,l,q_l-1}|) \cdot 2 > \epsilon \\
0 & \text{otherwise}
\end{cases},
\]
\end{minipage}
\end{center}

and a unique outlier vector is computed as \(\mathbf{o}_{l} = \mathbf{o}^d_l + \mathbf{o}^c_l\). When the current level is \(l=0\), all the previous operations are performed only on the detail components due to the absence of coarse decomposition.
Since our background assumptions are different from the standard statistical $t$-test, we prove that the Equation \eqref{eq:tscores} is correct. In particular, the statistical analysis available in Section \ref{formal_detivation_ttest} derives the ad-hoc $t$-score formula that follows a Student's $t$-distribution with \(q_l-1\) degrees of freedom. 

\subsection{Outlier Tree}

\begin{algorithm}
	\footnotesize
    \caption{BuildTree}
    \label{alg:build_tree}
    \begin{algorithmic}[1]
        \REQUIRE Time series size $M$, current level $l$, window $w$, max level $\hat{L}$, outlier vector $\mathbf{o}$, previous-level tree $\mathbf{z}_{p}$
        \ENSURE Current-level tree $\mathbf{z}$
        
        \STATE $\mathbf{z} = \mathbf{0}_{\frac{M}{2^l}}$ \COMMENT{Initialise the tree with zeros}
        
        \FOR{$i = 1$ \TO $\frac{M}{2^l} - w + 1$}
            \IF{$o_{i} = 1$}
                \STATE add $1$ to $\mathbf{z}[i : i+w-1]$ \COMMENT{Mark the outlier window}
            \ENDIF
        \ENDFOR
        
        \IF{$0 \leq l < \hat{L}$} 
            \FOR{$i = 1$ \TO $\frac{M}{2^l}$}
                \IF{$z_{i,p} > 0$} 
                    \STATE add $z_{i,p}$ to $\mathbf{z}[i \cdot 2 : (i \cdot 2) + 1]$ \COMMENT{Propagate from previous level}
                \ENDIF
            \ENDFOR
        \ENDIF
    \RETURN \(\mathbf{z}\)
    \end{algorithmic}
\end{algorithm}

The outlier vector \(\mathbf{o}_{l}\) contains information about the positions of the outliers at level \(l\). Since the length of the vector depends on the size of the window, the outlier positions can be extracted by building a tree as shown in Algorithm \ref{alg:build_tree}. 

\begin{figure*}
    \centering \includegraphics[width=0.90\textwidth]{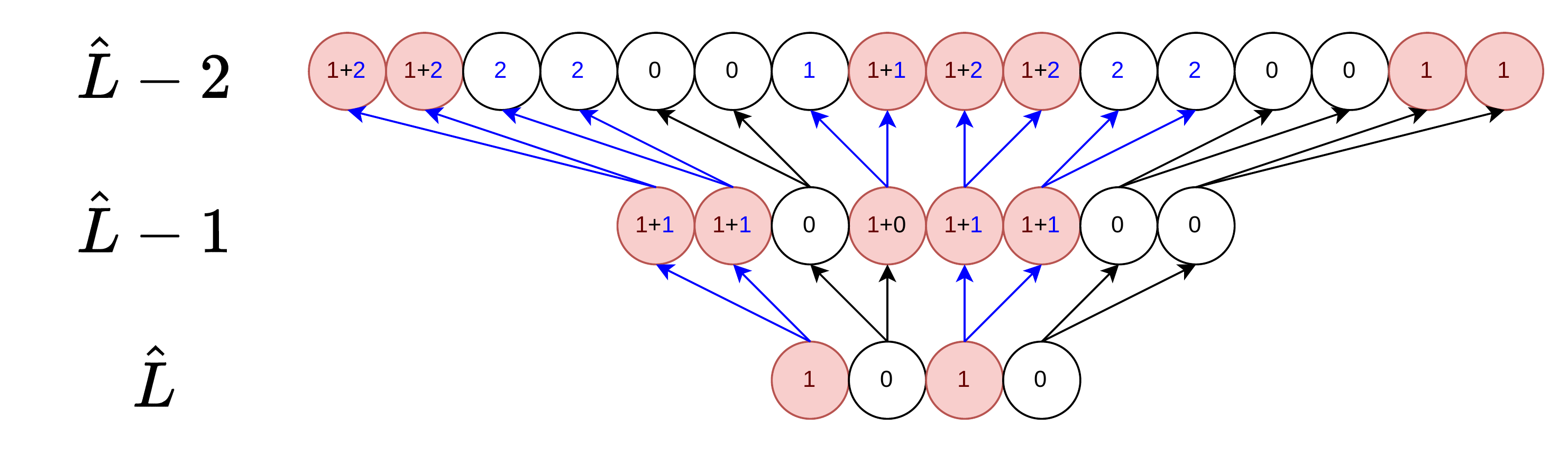}
    \caption{Example of outlier tree propagation. Starting from the lowest level \(\hat{L}\), the detected outliers contribute by setting to one the corresponding leaves (red circles). Level \(\hat{L}-1\) has the same detection behaviour of \(\hat{L}\), while retaining the information from the previous level (blue arrows).}
    \label{fig:tree}
\end{figure*}

The core logic of the latter is based on the assumption that windows containing outliers contribute more significantly to the cumulative sum during the construction of the tree (see Figure \ref{fig:tree}). In terms of notation, the slice operator $[a : b]$ follows the convention of being inclusive of the starting index $a$ and exclusive of the ending index $b$. At each iteration, the algorithm generates a vector $\mathbf{z}$ representing the tree at a specific decomposition level $l$. The dimension of $\mathbf{z}$ is dynamically adjusted to match the signal length at that stage, namely $\frac{M}{2^l}$, ensuring that outliers can be mapped back to their original positions within that specific temporal resolution.

To maintain computational efficiency and minimise memory overhead, the algorithm does not store all decomposition levels simultaneously. Instead, it proceeds iteratively by maintaining only the current-level tree $\mathbf{z}$ and the previous-level tree $\mathbf{z}_{p}$. If the current level $l$ satisfies the condition $0 \leq l < \hat{L}$, the algorithm propagates the anomaly information from the coarser level to the finer one. Specifically, for each point in the previous tree $\mathbf{z}_{p}$ with a non-zero score, the value is projected onto its corresponding children in the current vector $\mathbf{z}$. 

\begin{figure}
    \centering \includegraphics[width=0.48\textwidth]{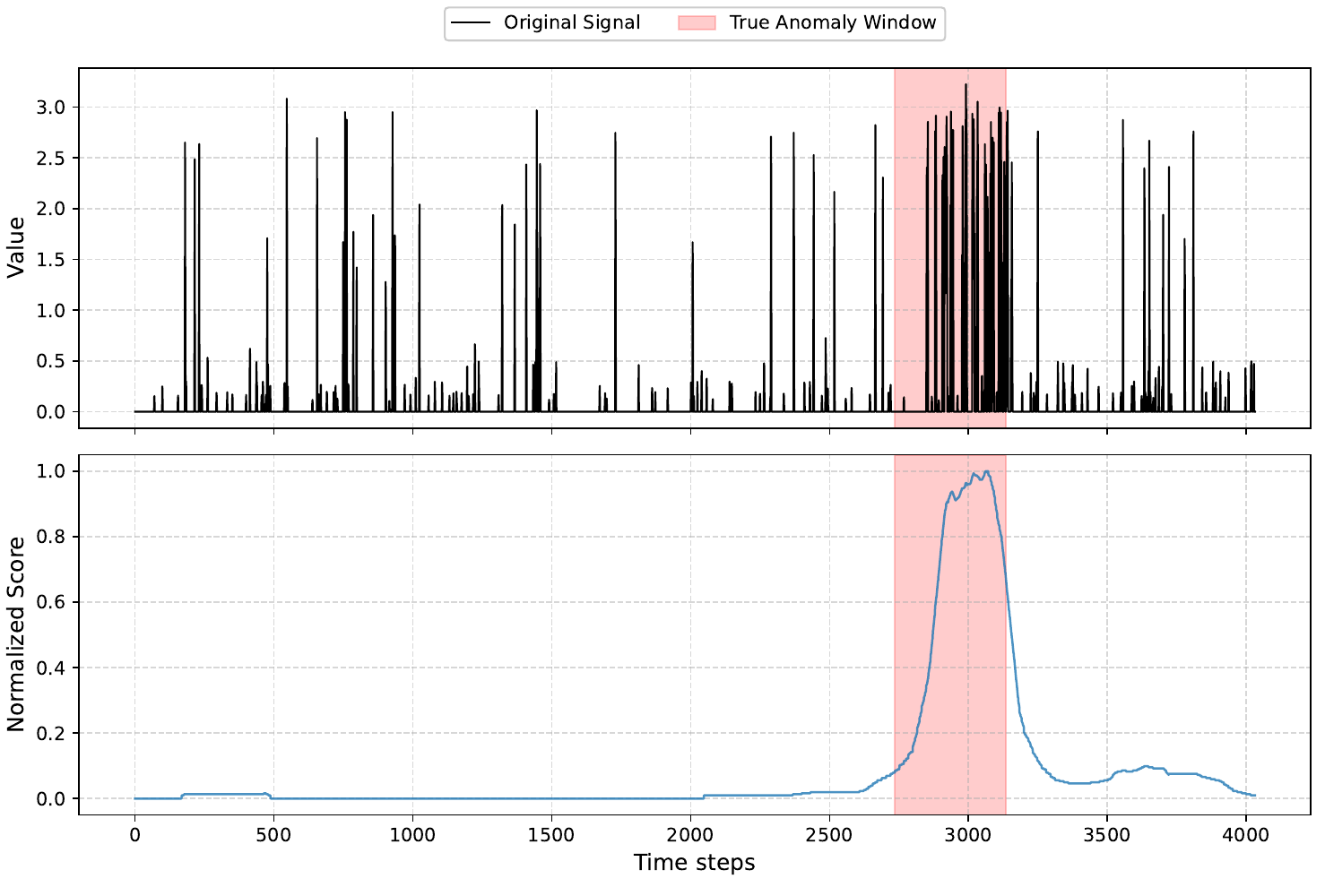}
    \caption{DWTt-test output scores for the “art\_load\_balancer\_spikes” dataset within the NAB collection.}
    \label{fig:dwtt-test_scores}
\end{figure}

By combining local insights across all $\hat{L}$ levels, the final output is an $M$-dimensional vector that exhibits peaks at the detected outlier locations (as illustrated in Figure \ref{fig:dwtt-test_scores}). 

Once the final outlier scores are computed, various thresholding or clustering techniques could be applied to isolate the anomalies. However, we deliberately avoid any further post-processing of the output scores. This choice is made to ensure a fair and transparent comparison with competing algorithms, utilising a ROC-based metrics that relies exclusively on the raw model output (further details are provided in Section \ref{results}).

\subsection{Formal derivation of the $t$-test}
\label{formal_detivation_ttest}

In this section, we present the formal derivation of the $t$-score, specifically tailored for the analysis of our algorithm. 
It is crucial to recognise that the classical Student's $t$-test is built upon a specific set of theoretical assumptions that are often violated in time series. Specifically, the standard derivation assumes that:

\begin{itemize}
    \item observations are independent and identically distributed (i.i.d.), such that $Cov(x_i, x_j) = 0$ for all $i \neq j$;
    \item the underlying population follows a normal distribution, $x_i \sim \mathcal{N}(\mu, \sigma^2)$;
    \item the parameters $\mu$ and $\sigma^2$ remain constant throughout the sampling process.
\end{itemize}

In the context of our algorithm, the resulting time series does not necessarily satisfy these conditions. Temporal dependencies, auto-correlation, or non-Gaussian noise distributions means that applying the classic $t$-test formula directly would yield biased results and unreliable $p$-values. 
Therefore, we show how to adapt the $t$-score computation to account for these characteristics, ensuring that the derived statistic maintains its theoretical properties and provides a valid measure of significance under non-ideal conditions.

\begin{figure}
    \centering \includegraphics[width=0.49\textwidth]{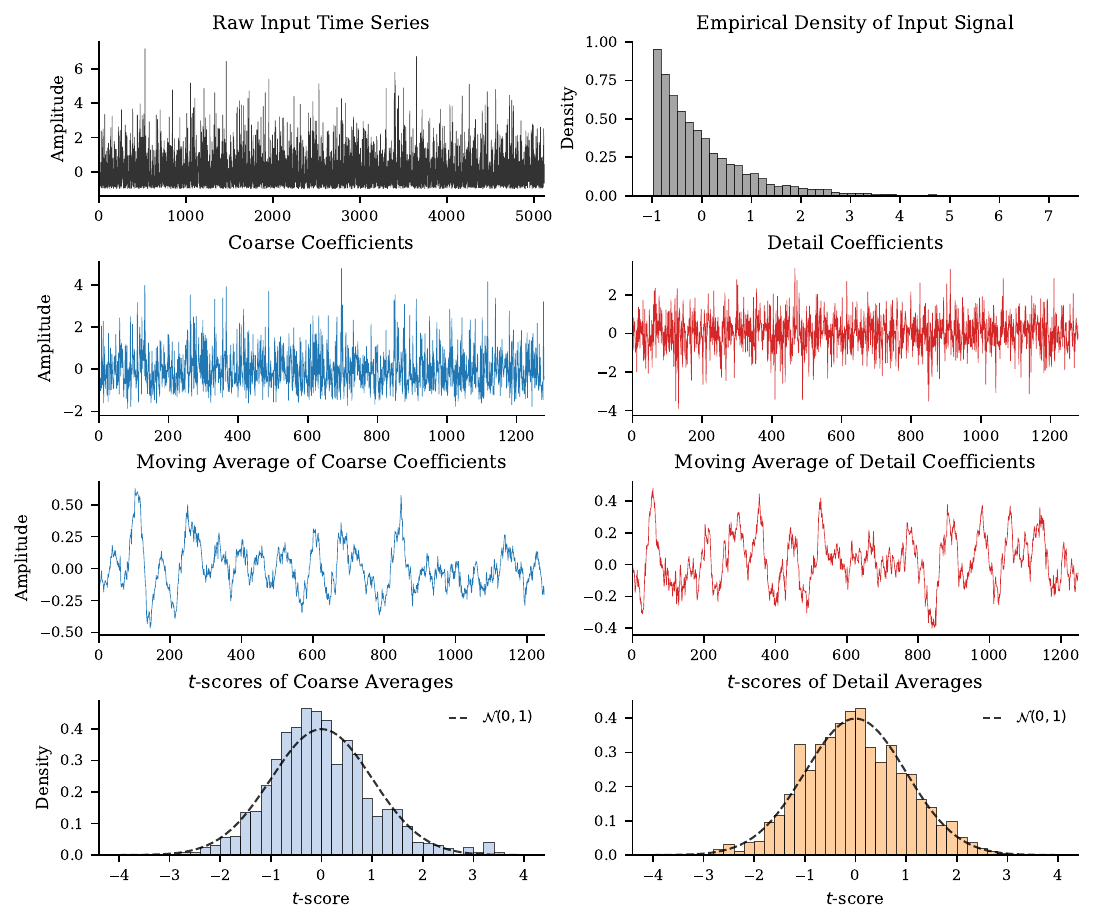}
    \caption{Statistical analysis of Haar wavelet coefficients for an i.i.d. time series. The raw input signal is constructed such that each observation $x_i$ is mutually independent and drawn from a target distribution. The plots illustrate the coarse and detail coefficients obtained at a specific decomposition level $l$. By applying a moving average with a sliding window of size $w$, we extract the local mean behaviour over time. The bottom panels empirically demonstrate the distribution of the resulting $t$-scores. Let $q_l$ be the number of elements in the resulting moving average vector; the $t$-scores follow a Student's $t$-distribution with $q_l-1$ degrees of freedom. Given the large sample size $q_l$, the $t$-distribution asymptotically converges and aligns with the Standard Normal distribution $\mathcal{N}(0,1)$ (solid black line).}
    \label{fig:wavelet_tscore}
\end{figure}

The derivation of a robust $t$-score formula relies on specific foundational assumptions. 
In particular, we assume that the observations of the original time series $x_i$ are 
independent and identically distributed (see Figure \ref{fig:wavelet_tscore}). 
While these assumptions simplify the inherent dependency of some time series, 
they represent a well-established standard in practical applications and exhibit 
remarkable reliability for outlier detection. To develop our ad-hoc $t$-score, 
we first formalise the requisite statistical framework.

\begin{definition}[Random Sample]
\label{def:random_sample}
Let $X = \{X_1, X_2, \dots, X_n\}$ be a set of $n$ independent and identically distributed (i.i.d.) random variables representing a univariate time series, such that $X_i \sim \mathcal{D}(\mu, \sigma^2)$, where $\mu$ is the population mean and $\sigma^2 < \infty$ is the population variance.
\end{definition}

\begin{definition}[Sample Statistics]
\label{def:sample_stats}
The Sample Mean $\bar{X}_n$ is defined as:
\[ \bar{X}_n = \frac{1}{n} \sum_{i=1}^{n} X_i. \]
The Sample Standard Deviation $S_n$ is defined using the unbiased sample variance:
\[ S_n = \sqrt{\frac{1}{n-1} \sum_{i=1}^{n} (X_i - \bar{X}_n)^2}. \]
\end{definition}

\begin{definition}[Sample Z-score and Studentisation]
\label{def:zscore}
Let $X = \{X_1, \dots, X_n\}$ be a random sample as defined in Definition \ref{def:random_sample}. We distinguish between two forms of score transformations:
\begin{enumerate}
    \item Population Z-score: If the population parameters $\mu$ and $\sigma$ are known, the Z-score for an observation $x_i$ is:
    \[ z_i = \frac{x_i - \mu}{\sigma}. \]
    \item Sample Z-score (Studentised score): If $\sigma$ or $\mu$ are unknown, they are replaced by the sample standard deviation $S_n$ and by the sample mean $\bar{X}_n$. In this case the transformation is termed \textit{studentisation}:
    \[ t_i = \frac{x_i - \bar{X}_n}{S_n}. \]
\end{enumerate}
\end{definition}

\begin{definition}[Chi-squared Distribution]
\label{def:chi_squared}
Let $Z_1, Z_2, \dots, Z_N$ be i.i.d. random variables such that $Z_i \sim \mathcal{N}(0, 1)$ for all $i = 1, \dots, N$. The random variable $K$ defined as:
\[ K = \sum_{i=1}^{N} Z_i^2, \]
follows a Chi-squared distribution with $N$ degrees of freedom, denoted as $K \sim \chi^2(N)$.
\end{definition}

\begin{definition}[$t$-Student Distribution]
\label{def:t_student}
Let $Z$ and $K$ be two independent random variables such that $Z \sim \mathcal{N}(0, 1)$ and $K \sim \chi^2(N)$. The random variable $T$ defined by the ratio:
\[ T = \frac{Z}{\sqrt{K/N}}, \]
follows a $t$-Student distribution with $N$ degrees of freedom, denoted as $T \sim t(N)$.
\end{definition}

\begin{theorem}[Central Limit Theorem]
\label{theo:clt}
Let $X_1, X_2, \dots, X_n$ be i.i.d. random variables from an arbitrary distribution $\mathcal{D}$ with mean $\mu$ and finite variance $\sigma^2$. As $n \to \infty$, the distribution of the standardised sample mean converges to the standard normal distribution:
\[ Z_n = \frac{\bar{X}_n - \mu}{\sigma / \sqrt{n}} \xrightarrow{d} \mathcal{N}(0, 1). \]
This implies that for large $n$, the sample mean is approximately distributed as $\bar{X}_n \sim \mathcal{N}(\mu, \frac{\sigma^2}{n})$.
\end{theorem}

\begin{theorem}[Convergence with Estimated Variance]
\label{theo:convergence}
In the context of the Central Limit Theorem (Theorem \ref{theo:clt}), if the population variance $\sigma^2$ is unknown and replaced by the unbiased sample variance $S_n^2$, the standardised statistic still converges in distribution to the standard normal distribution as $n \to \infty$ by Slutsky's Theorem:

\[ \frac{\bar{X}_n - \mu}{S_n / \sqrt{n}} \xrightarrow{d} \mathcal{N}(0, 1). \]

Furthermore, if the underlying random sample is drawn from a strictly Normal distribution ($X_i \sim \mathcal{N}(\mu, \sigma^2)$), this statistic follows an exact $t$-Student distribution with $n-1$ degrees of freedom for any sample size $n$:
\[ \frac{\bar{X}_n - \mu}{S_n / \sqrt{n}} \sim t(n-1). \]
\end{theorem}

Starting from these theoretical premises, and fixing the DWT level to $l=0$, we can explicitly compute the correct $t$-score for the proposed algorithm. We report here only the derivation for level $l=0$ since at this scale only details are available, streamlining the notation. Naturally, the analytical process remains identical for all other levels and for the coarse coefficients as well.

Given a time series formulated as in Definition \ref{def:random_sample} and an appropriate window size $w$, the first step of the DWTt-test is to compute the detail matrix $\mathbf{D}\in \mathbb{R}^{(M-w+1) \times w}$. To simplify the notation, let $q = M-w+1$. Next, the average of each row is computed, yielding the vector $\mathbf{d}^\text{avg} = \{d^\text{avg}_1, d^\text{avg}_2, \dots, d^\text{avg}_q \}$. According to Theorem \ref{theo:clt}, the distribution of each element can be approximated as $d^\text{avg}_i \sim \mathcal{N}(\mu,\frac{\sigma^2}{w})$. 

Relying on this normality property, we can compute the sample standard deviation $S_q$ of the vector $\mathbf{d}^\text{avg}$ following Definition \ref{def:sample_stats}. Then, following the first case of Definition \ref{def:zscore}, the population Z-score for the $i$-th element is defined as:
\[ Z_i = \frac{d^\text{avg}_i - \mu}{\sigma / \sqrt{w}}. \]
Since each $d^\text{avg}_i \sim \mathcal{N}\left(\mu, \frac{\sigma^2}{w}\right)$, it trivially follows that the standardised variable is distributed as a standard normal, i.e., $Z_i \sim \mathcal{N}(0, 1)$.

To fulfill the requirements of the Student's $t$-distribution as stated in Definition \ref{def:t_student}, we must identify a random variable $K$ that follows a Chi-squared distribution (Definition \ref{def:chi_squared}). Let us analyse the sum of the squared standardised deviations of our details vector:

\[ \sum_{j=1}^{q} \frac{(d^\text{avg}_j - \mu)^2}{\sigma^2/w} \sim \chi^2(q). \]

This relationship perfectly aligns with Definition \eqref{def:chi_squared} because each term inside the sum is a squared standard normal variable, $\left(\mathcal{N}(0,1)\right)^2 \sim \chi^2(1)$. By partitioning this sum of squares, we can rewrite the equation as:

\[ \underbrace{\sum_{j=1}^{q} \frac{(d^\text{avg}_j - \mu)^2}{\sigma^2/w}}_{\sim \chi^2(q)} = \underbrace{\frac{(q-1)S_q^2 w}{\sigma^2}}_{K} + \underbrace{\frac{q(\bar{d}^\text{avg} - \mu)^2 w}{\sigma^2}}_{\sim \chi^2(1)}, \]

where $S_q^2$ is the sample variance of $\mathbf{d}^\text{avg}$ and $\bar{d}^\text{avg}$ is its sample mean. Note that $\bar{d}^\text{avg} = \frac{1}{q} \sum_{j=1}^q d^\text{avg}_j \sim \mathcal{N}\left(\mu, \frac{\sigma^2}{w \cdot q}\right)$.

Since the left-hand side follows a $\chi^2(q)$ distribution and the rightmost term follows a $\chi^2(1)$ distribution, the remaining term directly yields our variable $K$:

\[ K = \frac{w(q-1)S_q^2}{\sigma^2} \sim \chi^2(q-1). \]

Now we can derive a valid $t$-score. According to Definition \ref{def:t_student}, the $t$-statistic with $q-1$ degrees of freedom is constructed via the ratio of the standard normal variable $Z_i$ and $\sqrt{K/(q-1)}$. For a single observation $d^\text{avg}_i$, substituting our derived $Z_i$ and $K$, the studentised score (consistent with Definition \ref{def:zscore}, case 2) evaluates to:

\begin{equation}
t_{q-1} = \frac{d^\text{avg}_i - \mu}{S_q}.
\end{equation}

Remarkably, the resulting formulation is algebraically independent of the window size $w$. To formally prove that this specific equation correctly yields a Student's $t$-distribution, we expand the denominator by multiplying and dividing by the necessary variance components:

\begin{equation}
\begin{split}
t_{q-1} &= \frac{d^\text{avg}_i - \mu}{\sqrt{S_q^2}} \\
&= \frac{d^\text{avg}_i - \mu}{\sqrt{S_q^2 \frac{(\sigma^2/w)(q-1)}{(\sigma^2/w)(q-1)}}} \\
&= \frac{d^\text{avg}_i - \mu}{\sqrt{\frac{\sigma^2}{w}} \cdot \sqrt{\frac{w(q-1)S_q^2}{\sigma^2}} \cdot \sqrt{\frac{1}{q-1}}} \\
&= \frac{\frac{d^\text{avg}_i - \mu}{\sigma/\sqrt{w}}}{\sqrt{\frac{\frac{w(q-1)S_q^2}{\sigma^2}}{q-1}}} \\
&= \frac{Z_i}{\sqrt{K / (q-1)}}
\end{split}
\end{equation}

This concludes the proof. In accordance with Theorem \ref{theo:convergence} we have proved that, despite the initial variance scaling induced by the parameter $w$, the derived statistic satisfies the theoretical requirements (Definition \ref{def:t_student}) and guarantees an exact Student's $t$-distribution, $t_{q-1} \sim t(q-1)$.

\subsection{Computational Complexity Analysis}
\label{sec:complexity}

In this section, we analyse the computational complexity of the proposed DWTt-test algorithm (Algorithm 1). Let $N$ be the dimension of the input time series $\mathbf{x}$. The algorithm first pads the input to the next power of two, $M = 2^{\lceil \log_2(N) \rceil}$. Since $M < 2N$, we have that $M = \mathcal{O}(N)$. We evaluate the complexity by analysing the main algorithmic steps.

\begin{enumerate}
    \item DWT: Computing the Haar wavelet transform up to a maximum level $\hat{L}$ requires processing signals of halving sizes at each step. The size of the detail and approximation coefficients at level $l$ is $\frac{M}{2^l}$. The total number of operations for the DWT across all $\hat{L}$ levels is bounded by the geometric series:
    \[ \sum_{l=1}^{\hat{L}} \mathcal{O}\left(\frac{M}{2^l}\right) = \mathcal{O}\left( M \sum_{l=1}^{\hat{L}} \frac{1}{2^l} \right) \le \mathcal{O}(M) = \mathcal{O}(N). \]
    Thus, the wavelet decomposition takes strictly linear time.

    \item Sliding Window and Statistics Computation: At each level $l \in [1, \hat{L}]$, a sliding window of size $w_l = w(\hat{L} - l + 1)$ is applied. The length of the signal at level $l$ is $\frac{M}{2^l}$, meaning the number of windows generated at this level is $q_l = \frac{M}{2^l} - w_l + 1$. 
    
    To determine the overall complexity, we sum the number of windows across all $\hat{L}$ levels. Notice that the window size $w_l$ scales linearly. By substituting $j = \hat{L} - l + 1$, the sum of the window sizes becomes an arithmetic progression:
    
    \begin{equation}
    \begin{split}
    \sum_{l=1}^{\hat{L}} q_l &= \sum_{l=1}^{\hat{L}} \left( \frac{M}{2^l} - w(\hat{L} - l + 1) + 1 \right) \\
    &= \sum_{l=1}^{\hat{L}} \frac{M}{2^l} - w \sum_{j=1}^{\hat{L}} j + \sum_{l=1}^{\hat{L}} 1 \\
    &\le M - w \frac{\hat{L}(\hat{L}+1)}{2} + \hat{L}.
    \end{split}
    \end{equation}
    
    Since $\hat{L}$ (the maximum decomposition level, typically 3 or 4) and the base window parameter $w$ are empirically chosen constants independent of $N$, the terms involving them evaluate to $\mathcal{O}(1)$ with respect to the input size. Therefore, the total number of windows processed across the entire hierarchy is $\mathcal{O}(M) - \mathcal{O}(\hat{L}^2) + \mathcal{O}(\hat{L}) = \mathcal{O}(N)$. 
    
    The computation of the sample mean and variance for each window can be optimised to $\mathcal{O}(1)$ per window using cumulative sums, or at worst $\mathcal{O}(w \hat{L})$ naively. Since $w \hat{L}$ is a small constant, the overall sliding window processing time remains $\mathcal{O}(N)$.

    \item Score Evaluation and Tree Building: For each of the $\mathcal{O}(N)$ total windows, computing the $t$-score, evaluating the $p$-value against the threshold $\epsilon$, and checking the anomaly condition requires a constant $\mathcal{O}(1)$ number of operations. Consequently, the scoring mechanism takes $\mathcal{O}(N)$ time. Similarly, updating the hierarchical anomaly tree requires processing elements progressively up the levels, which can be stated again as the geometric sum $\sum \frac{M}{2^l} \le M$, bounding the tree operations to $\mathcal{O}(N)$.

    \item DWT $l=0$ Analysis: Finally, the algorithm performs a similar sliding window pass over the original padded series of size $M$ (level $l=0$). This generates $M - w + 1$ windows. Processing these windows and assigning the final scores takes $\mathcal{O}(M) = \mathcal{O}(N)$ time.
\end{enumerate}

Summing the linear complexities of the individual stages, the overall time complexity of the DWTt-test algorithm is strictly \textbf{$\mathcal{O}(N)$}. This linear scalability guarantees extreme efficiency, making the algorithm highly suitable for real-time anomaly detection in large univariate time series.
\label{computational_complexity}

\section{Experimental Results}
\label{results}
In this section, we present the comprehensive evaluation of the proposed anomaly detection algorithm. To ensure a fair and rigorous comparison, we strictly adhere to a threshold-agnostic evaluation protocol, assessing the raw anomaly scores produced by the models rather than binary labels.

\subsection{Evaluation Methodology}

In many recent time series anomaly detection studies, particularly those focusing on self-supervised deep learning models, the evaluation is heavily biased by the use of \textit{point adjustment} (PA) protocols \cite{dcdetector, SELLAM2025, Xu2022}. Under PA, if a single anomalous point within a true anomalous segment is detected, the entire segment is considered correctly predicted. This significantly inflates the apparent performance of the algorithms and masks their actual detection capabilities. 

To overcome this flaw, we follow the methodology proposed by \cite{Schmidl2022}, removing any thresholding that transforms continuous anomaly scores into binary labels. The translation of anomaly scores into binary anomaly labels via thresholding is an orthogonal, algorithm-independent problem that we do not consider in this evaluation. By relying solely on the raw continuous scorings, we evaluate the intrinsic discriminative power of the models.

\subsection{Datasets and Experimental Setup}

Our evaluation is conducted exclusively on univariate time series data. We selected a diverse set of well-known benchmark datasets, encompassing both real-world recordings and synthetic series.
These datasets have been preprocessed and made available by \cite{Schmidl2022}. A summary of the datasets used in our experiments is provided in Table \ref{tab:datasets}.
The experimental setup differs based on the learning paradigm of the evaluated algorithms:
\begin{itemize}
    \item Unsupervised Algorithms: These models do not rely on prior training. To ensure optimal performance without data leakage, their hyper-parameters were globally tuned via a grid search exclusively on the GutenTAG dataset. We utilised the Good Time Series Anomaly Generator (GutenTAG) because it provides time series with a large variety of well-labeled anomalies. After tuning, the unsupervised models were tested across all the datasets listed in Table \ref{tab:datasets} except GutenTAG;
    \item Self-supervised Algorithms: These models require distinct training and testing phases. Therefore, they were trained and evaluated exclusively on the NASA-SMAP and NASA-MSL collections. These two macro-datasets are inherently partitioned into explicit training and test subsets, making them perfectly suited for evaluating supervised methods without introducing bias.
\end{itemize}
\begin{table}
\centering
\caption{Summary of the univariate datasets used for the experimental evaluation.}
\label{tab:datasets}
\resizebox{\columnwidth}{!}{%
\begin{tabular}{llrr}
\hline
\textbf{Dataset} & \textbf{Origin} & \textbf{Avg. Length} & \textbf{\# Datasets} \\ \hline
GutenTAG \cite{Schmidl2022} & Synthetic & 10,000 & 193 \\
NASA-SMAP \cite{nasa} & Real & 8,070 & 54 \\
NASA-MSL \cite{nasa} & Real & 2,730 & 27 \\
NAB \cite{nab} & Real/Synthetic & 6,302 & 58 \\
MGAB \cite{mgab} & Synthetic & 100,000 & 10 \\
Dodgers \cite{dodgers} & Real & 50,400 & 1 \\ \hline
\end{tabular}%
}
\end{table}

\subsection{Competing Algorithms and Baselines}
\label{sec:competitors}

\begin{table}
\centering
\caption{Summary of the competing algorithms used as baselines in our evaluation.}
\label{tab:competitors}
\begin{tabular}{ll}
\hline
\textbf{Algorithm} & \textbf{Method Family} \\ \hline
\multicolumn{2}{c}{\textit{Traditional Unsupervised Baselines}} \\ \hline
DWT-MLEAD \cite{dwtmlead} & Distribution \\
Left STAMPI \cite{lstampi} & Distance \\
MedianMethod \cite{medianmethod} & Forecasting \\
PCI \cite{pci}  & Reconstruction \\
Sub-IF \cite{sif} & Trees \\
Sub-LOF \cite{sLOF}  & Distance \\ \hline
\multicolumn{2}{c}{\textit{Deep Learning self-supervised Baselines}} \\ \hline
Anomaly Transformer \cite{Xu2022}  & Transformer \\
DCdetector \cite{dcdetector}  & Dual Attention \\
Mamba \cite{SELLAM2025} & State Space Model \\ \hline
\end{tabular}
\end{table}

To comprehensively evaluate the performance of our proposed method, we compare it against a diverse and robust set of state-of-the-art anomaly detection algorithms. A summary of the selected competing models, their learning paradigm, and the sources of their implementations is provided in Table \ref{tab:competitors}.
The selected baselines are strategically divided into two categories to cover both traditional, parameter-efficient methods and recent deep learning advancements: Unsupervised algorithms and self-supervised models.

For the unsupervised category, we selected several algorithms identified in the extensive benchmark conducted by Schmidl et al. \cite{Schmidl2022}. This selection ensures that our approach is compared against the most competitive traditional methods available in the literature. The chosen unsupervised competing algorithms include DWT-MLEAD \cite{dwtmlead}, Left STAMPI \cite{lstampi}, MedianMethod (MM) \cite{medianmethod}, PCI \cite{pci}, Subsequence Isolation Forest (Sub-IF) \cite{sif}, and Subsequence Local Outlier Factor (Sub-LOF) \cite{sLOF}. To guarantee maximum reproducibility, fairness, and consistency in our evaluation, we utilise the exact implementations and wrappers provided by the authors of the benchmark \cite{Schmidl2022}.

For the self-supervised category, we selected three recent, highly influential deep learning architectures that represent the current state-of-the-art in time series anomaly detection: Anomaly Transformer \cite{Xu2022}, DCdetector \cite{dcdetector}, and a Mamba-based architecture \cite{SELLAM2025}. It is worth noting here that these models are self-supervised rather than supervised, owing to the fact that they are trained ignoring the labels, and generate a supervisory signal in the form of a score that guides their loss function minimisation.

For all the models under test, we rely on the official open-source implementations provided by the respective authors to ensure their architectures are evaluated exactly as originally proposed. In particular, regarding the implementations of unsupervised methods, all them have been developed using Python. Therefore, the comparison is fair with regard to the actual running time. However, self-supervised models rely on the use of GPUs for their training and inference. As a consequence, their running time is not directly comparable to that of the unsupervised models.

\subsection{Evaluation Metrics}

To evaluate the various scorings, we use three threshold-agnostic Area Under the Curve (AUC) measures: The Area under the Receiver Operating Characteristics Curve (AUC-ROC), the Area Under the Precision-Recall Curve (AUC-PR), and the Area Under the range-based Precision, range-based Recall Curve (AUC-$P_T R_T$) \cite{ptpr}. These metrics provide different perspectives on the model's performance, ranging from overly optimistic to highly rigorous.

\subsubsection{AUC-ROC}
The ROC curve plots the True Positive Rate ($\mathrm{TPR}$) against the False Positive Rate ($\mathrm{FPR}$) across all possible thresholds:
\[ \mathrm{TPR} = \frac{\mathrm{TP}}{\mathrm{TP + FN}}, \quad \mathrm{FPR} = \frac{\mathrm{FP}}{\mathrm{FP + TN}}, \]
where $\mathrm{TP}$ are the True Positives, $\mathrm{FN}$ are the False Negatives, $\mathrm{FP}$ are the False Positives, and $\mathrm{TN}$ are the True Negatives. While AUC-ROC is widely used and serves to relate our results to existing evaluation results, it can be misleading in time series anomaly detection. Because anomalies are exceptionally rare, the number of negative samples ($\mathrm{N = FP + TN}$) is extremely large. A higher $\mathrm{N}$ is potentially reflected in both $\mathrm{FP}$ and $\mathrm{N}$, meaning the $\mathrm{FPR}$ ($\frac{\mathrm{FP}}{\mathrm{N}}$) is less influenced by $\mathrm{N}$ and remains artificially low even if the model predicts many false positives. Consequently, AUC-ROC tends to reward sensitive algorithms but often paints an overly optimistic picture.

\subsubsection{AUC-PR}
The Precision-Recall curve plots Precision against Recall ($\mathrm{TPR}$):
\[ \text{Precision} = \frac{\mathrm{TP}}{\mathrm{TP + FP}}. \]
Unlike $\mathrm{FPR}$, Precision is highly sensitive to class imbalance. The number of negative samples ($\mathrm{N}$) in a dataset has a higher influence on AUC-PR than on AUC-ROC since the precision is the ratio of true positives to all predicted positives. A higher $\mathrm{N}$ introduces more potentially false positives and, thus, lowers the Precision. Therefore, AUC-PR is a much harder metric that strictly rewards precise algorithms capable of minimising false alarms in highly imbalanced time series.

\subsubsection{AUC-$P_T R_T$}
While AUC-PR is robust, its point-wise nature can be excessively punitive for subsequence anomalies. To address this, we also employ range-based Precision ($P_T$) and range-based Recall ($R_T$), which are more recent evaluation metrics specifically developed for time series anomalies \cite{ptpr}. AUC-$P_T R_T$ simply calculates the area under the curve for these two new metrics. This metric softens the very strict preciseness requirements of AUC-PR to adapt the measure to subsequences, allowing for a more contextual evaluation of anomalous segments.

\subsection{Unsupervised Algorithms Evaluation}
\begin{figure}     
	 \centering
     \begin{subfigure}{\columnwidth}
         \centering
         \includegraphics[width=0.90\textwidth]{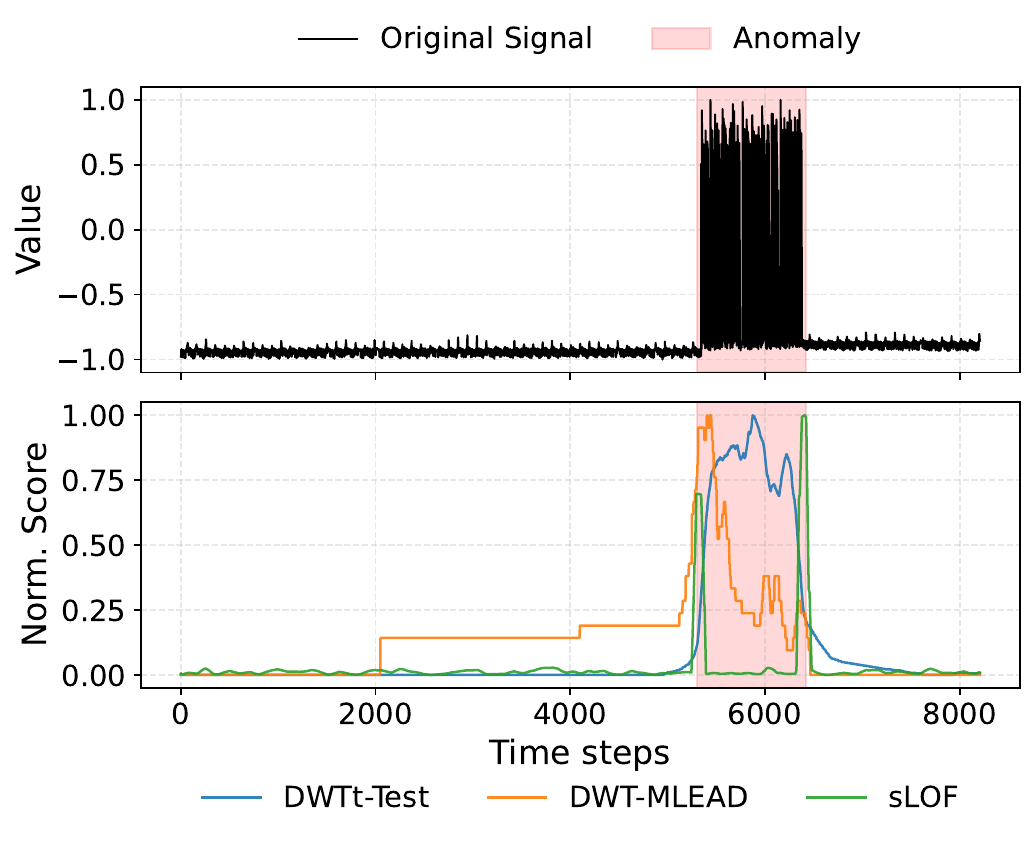}
         \caption{Anomaly scores for the P-2 dataset (SMAP collection)}
         \label{fig:SMAP_P2_UNS}
     \end{subfigure}

     \vspace{0.3cm}

     \begin{subfigure}{\columnwidth}
         \centering
         \includegraphics[width=0.90\textwidth]{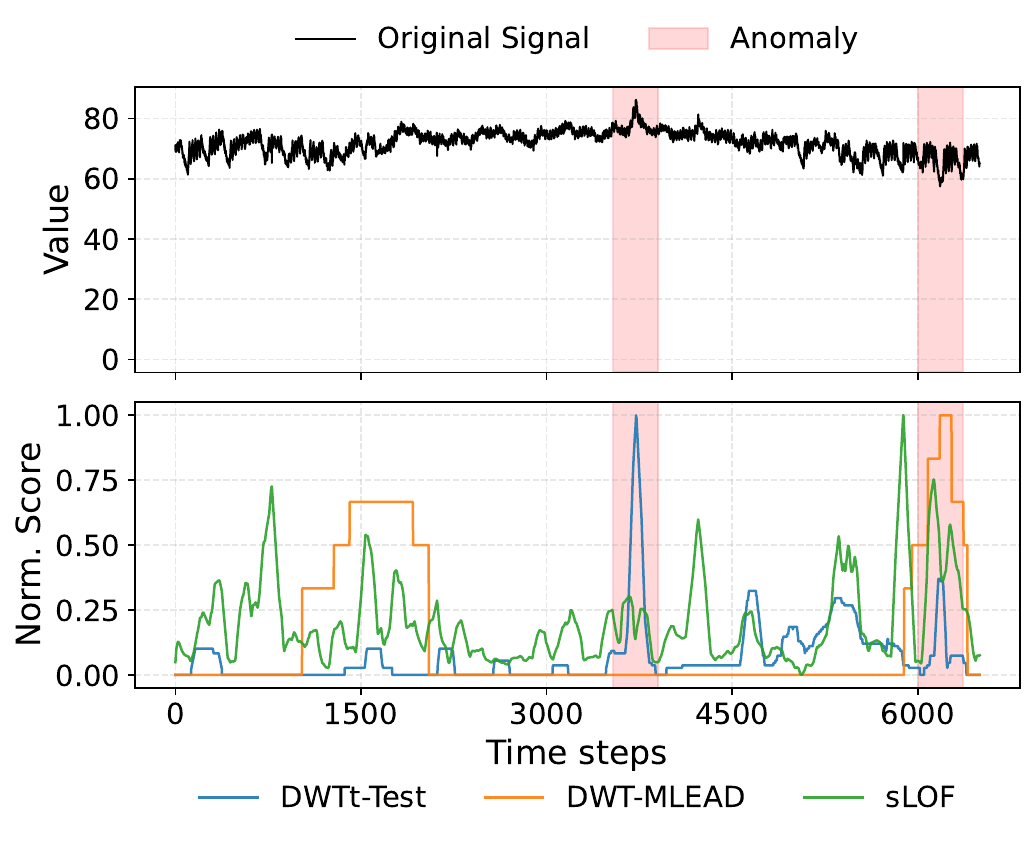}
         \caption{Anomaly scores for temperature failure (NAB collection)}
         \label{fig:NAB_AMBIENT_UNS}
     \end{subfigure}
     
     \caption{Comparison of unsupervised anomaly scores generated by DWTt-test, DWT-MLEAD, and sLOF. Each subplot illustrates the detection scores overlaid on the original time series for specific datasets from the SMAP and NAB collections.}
     \label{fig:combined_plots_uns}
\end{figure}
\begin{figure}
     \centering
     \ContinuedFloat
     
     \begin{subfigure}{\columnwidth}
         \centering
         \includegraphics[width=0.90\textwidth]{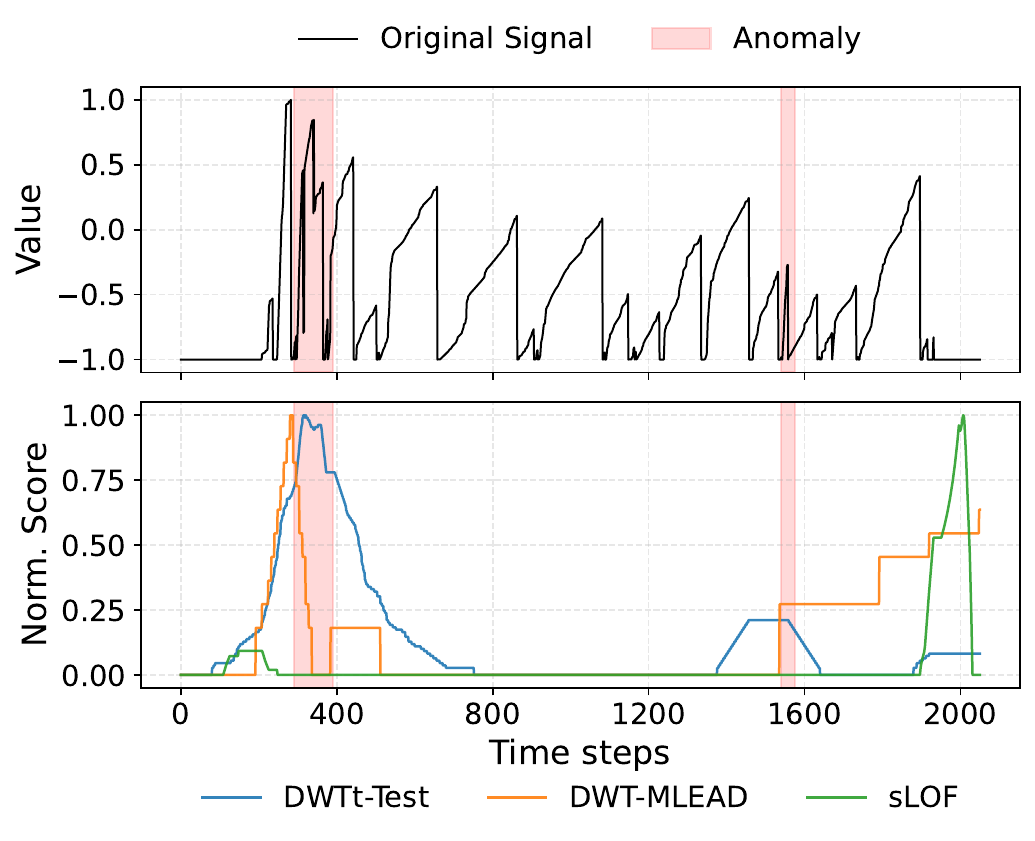}
         \caption{Anomaly scores for the C-2 dataset (MSL collection)}
         \label{fig:MSL_C2_UNS}
     \end{subfigure}

     \vspace{0.3cm}

     \begin{subfigure}{\columnwidth}
         \centering
         \includegraphics[width=0.90\textwidth]{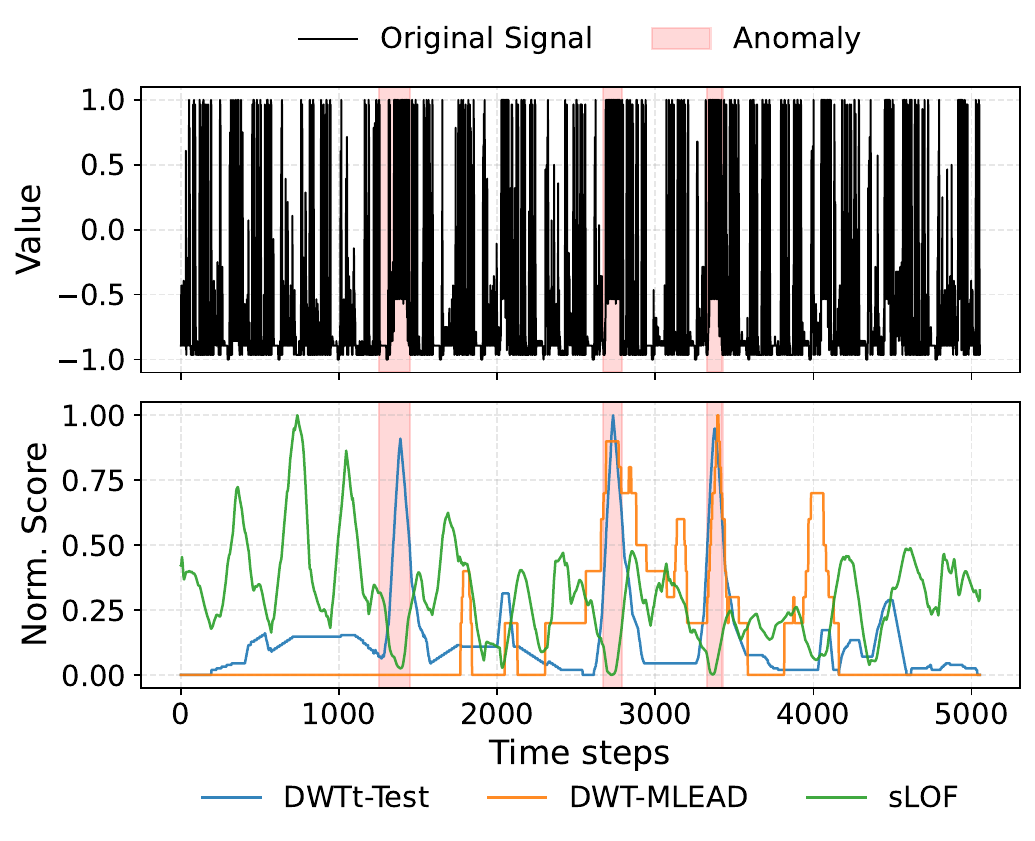}
         \caption{Anomaly scores for the F-7 dataset (MSL collection)}
         \label{fig:MSL_F7_UNS}
     \end{subfigure}
     
     \caption{Comparison of unsupervised anomaly scores generated by DWTt-test, DWT-MLEAD, and sLOF. Each subplot illustrates the detection scores overlaid on the original time series for specific datasets from the MSL collection.}
\end{figure}

\begin{figure}
     \centering
     \begin{subfigure}{\columnwidth} 
         \centering
         \includegraphics[width=0.95\textwidth]{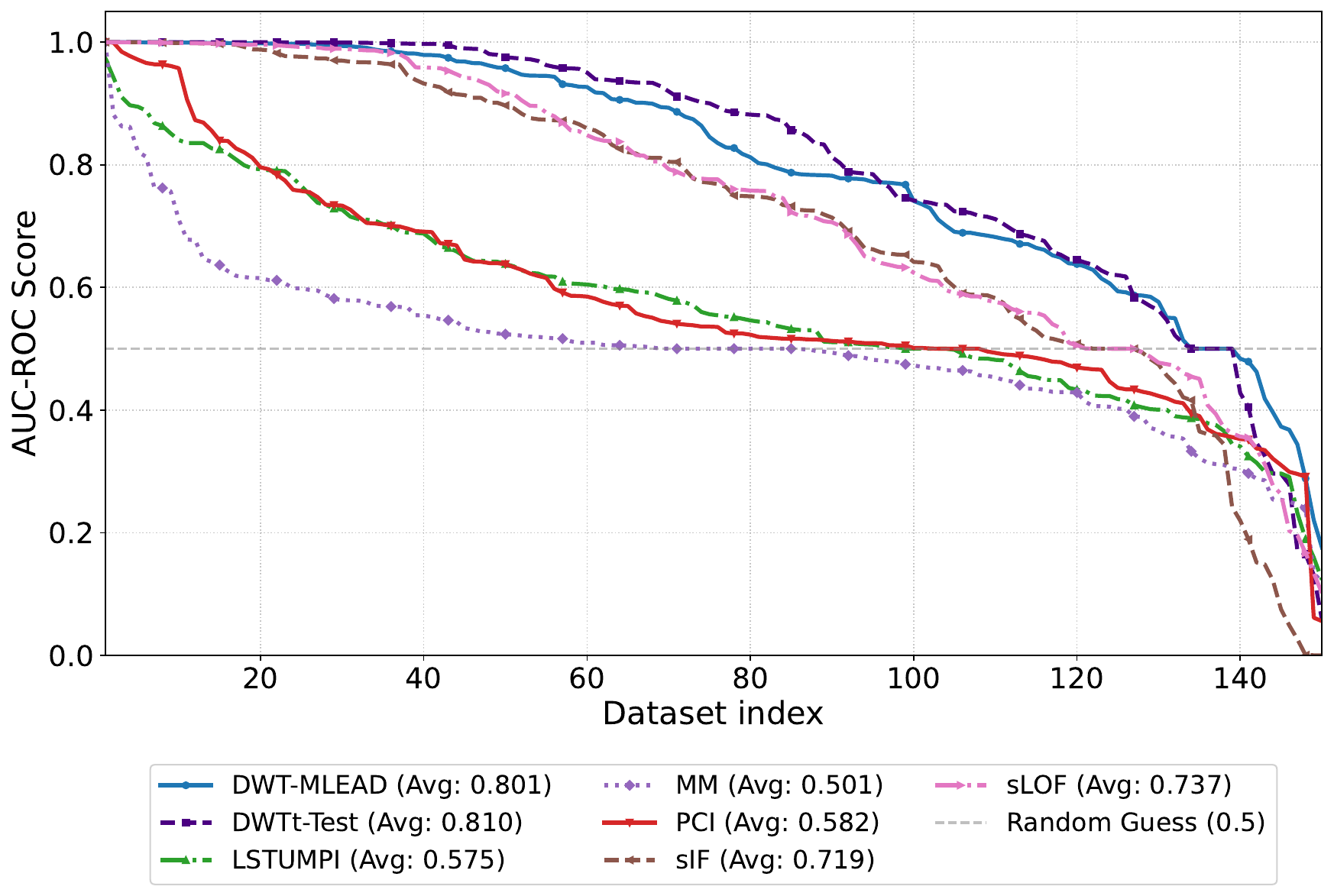}
         \caption{AUC-ROC values across all test datasets.}
         \label{fig:ROC_UNS_CURVE}
     \end{subfigure}

     \begin{subfigure}{\columnwidth}
         \centering
         \includegraphics[width=0.95\textwidth]{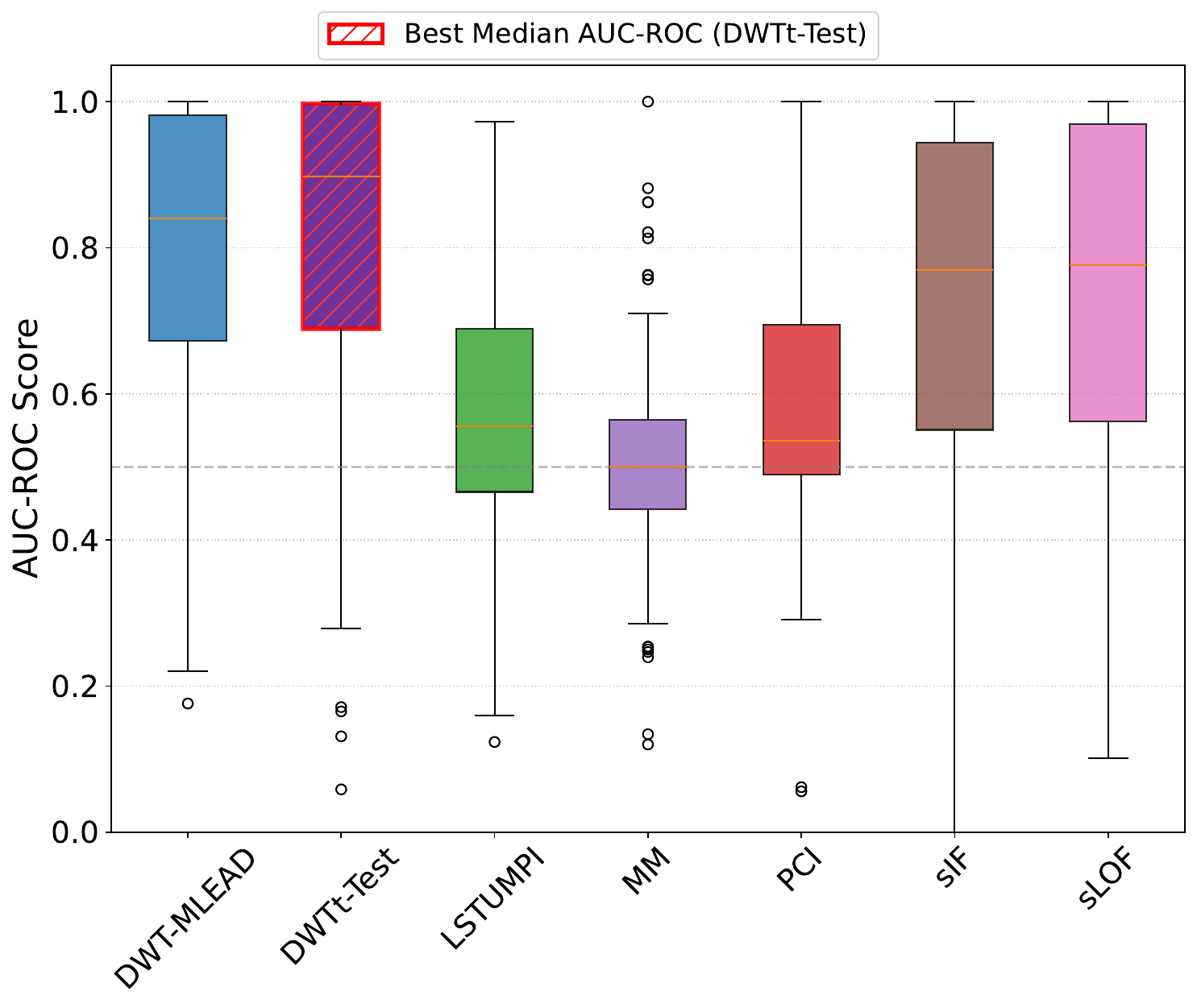}
         \caption{Distribution of AUC-ROC scores.}
         \label{fig:ROC_UNS_BOX}
     \end{subfigure}
     
     \caption{Quantitative evaluation of unsupervised algorithms based on AUC-ROC. The line plot (a) shows the metric value achieved for each dataset index, while the boxplot (b) provides a statistical summary of the score distribution across the entire collection.}
     \label{fig:ROC_UNS}
\end{figure}

\begin{figure}
     \centering
     \begin{subfigure}{\columnwidth}
         \centering
         \includegraphics[width=0.95\textwidth]{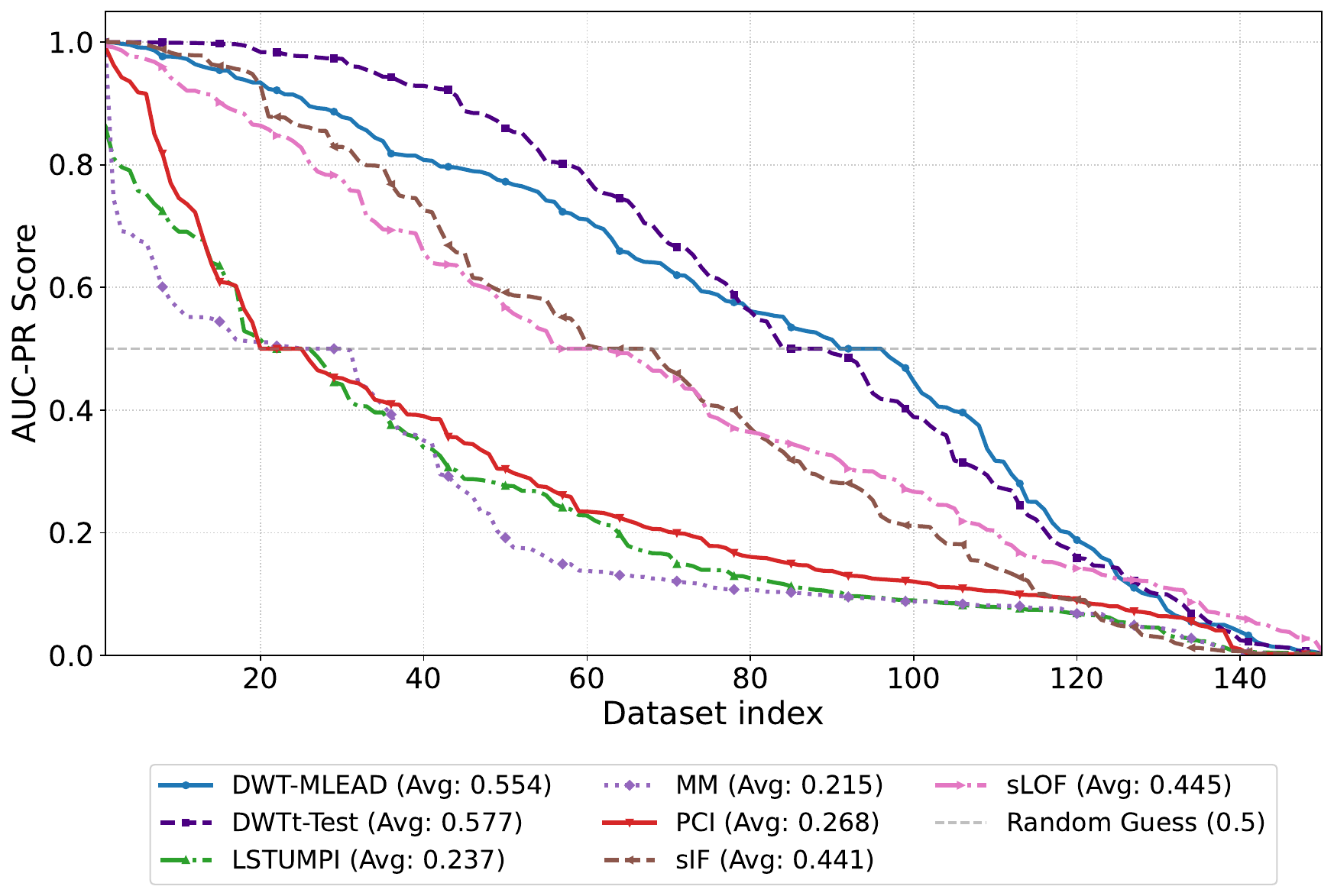}
         \caption{AUC-PR values across all test datasets.}
         \label{fig:PR_UNS_CURVE}
     \end{subfigure}
     
     \begin{subfigure}{\columnwidth}
         \centering
         \includegraphics[width=0.95\textwidth]{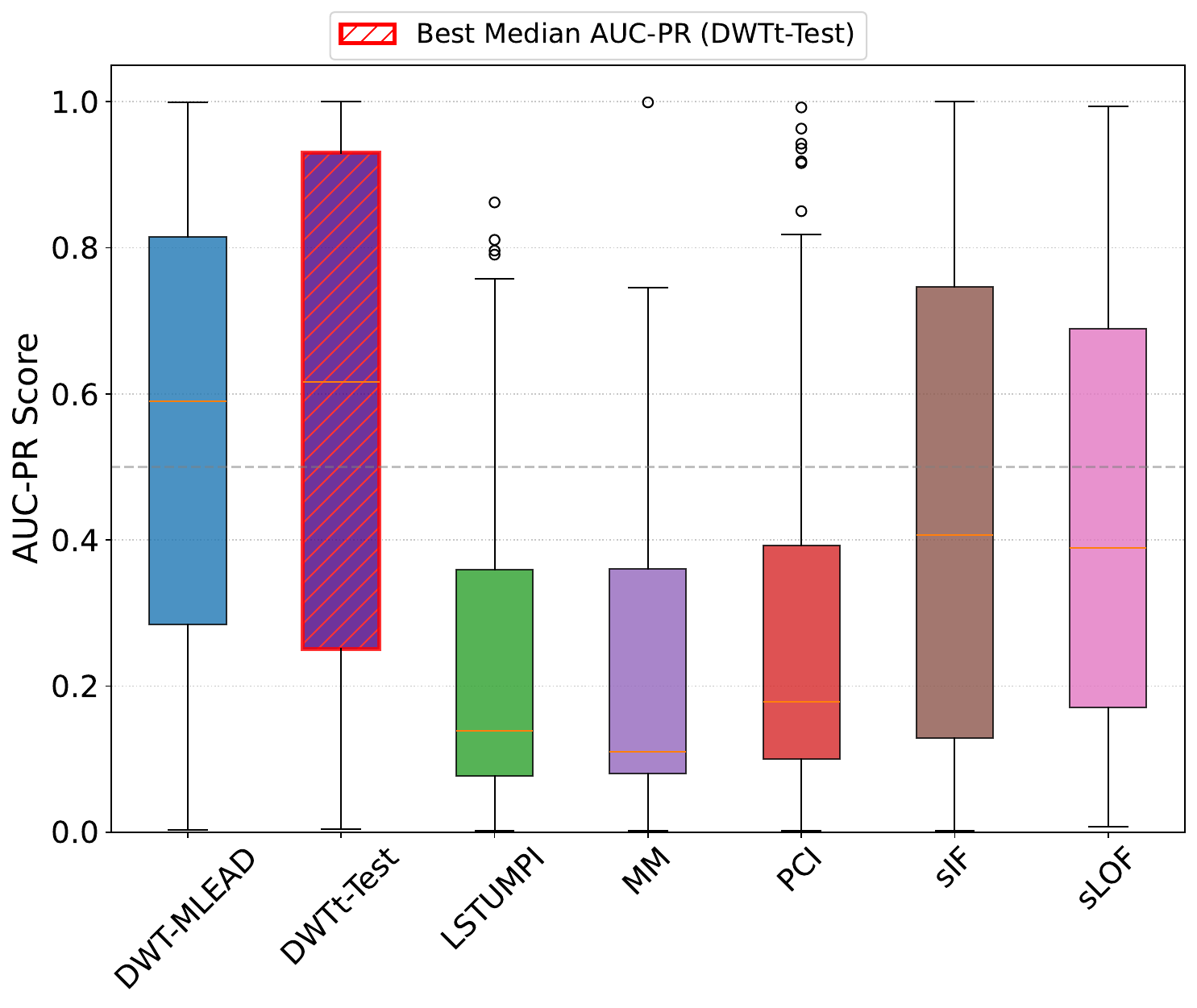}
         \caption{Distribution of AUC-PR scores.}
         \label{fig:PR_UNS_BOX}
     \end{subfigure}
     
     \caption{Quantitative evaluation of unsupervised algorithms based on AUC-PR. The line plot (a) shows the metric value achieved for each dataset index, while the boxplot (b) provides a statistical summary of the score distribution across the entire collection.}
     \label{fig:PR_UNS}
\end{figure}

\begin{figure}
     \centering
     \begin{subfigure}{\columnwidth}
         \centering
         \includegraphics[width=0.95\textwidth]{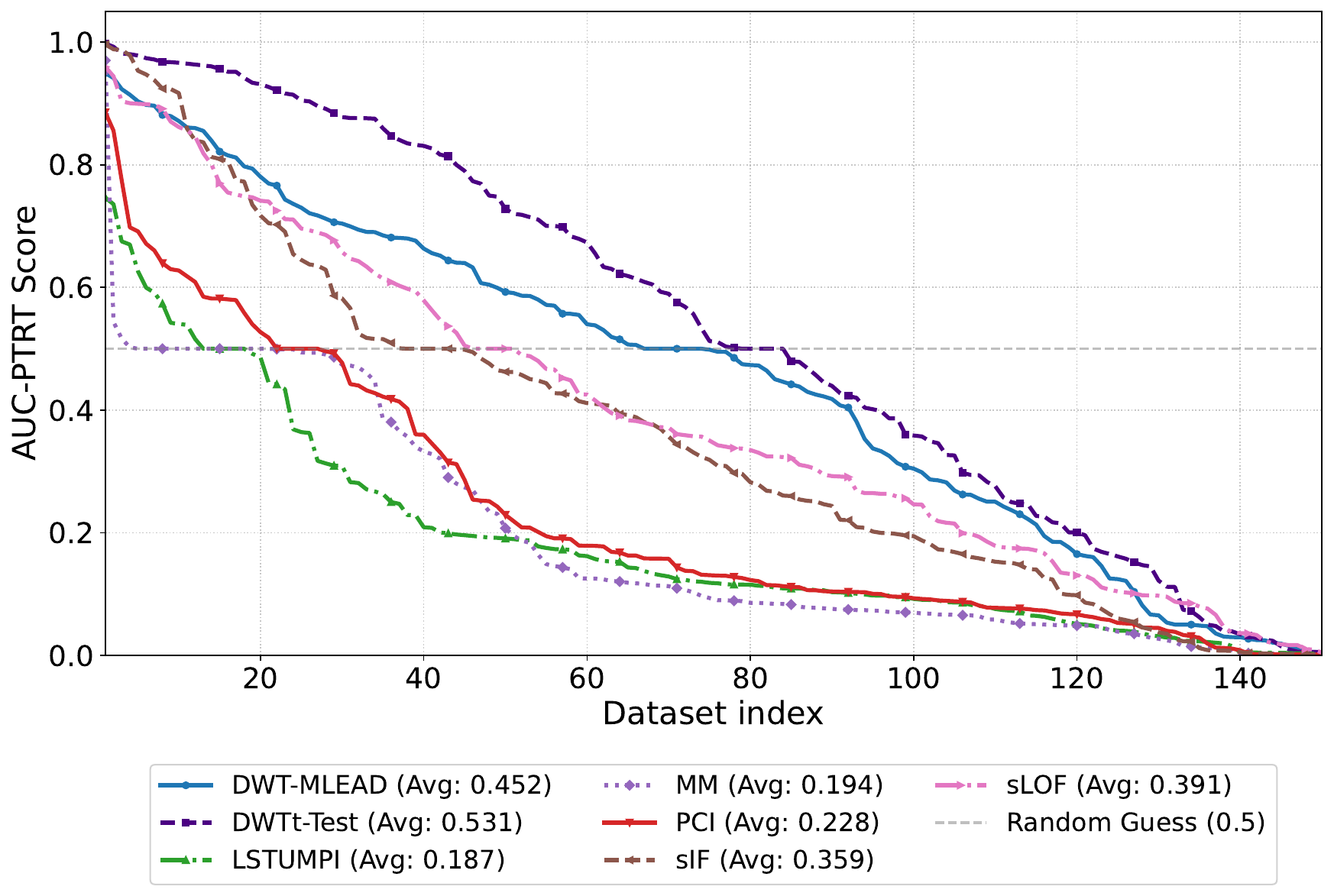}
         \caption{AUC-PTRT values across all test datasets.}
         \label{fig:PTRT_UNS_CURVE}
     \end{subfigure}
     
     \begin{subfigure}{\columnwidth}
         \centering
         \includegraphics[width=0.95\textwidth]{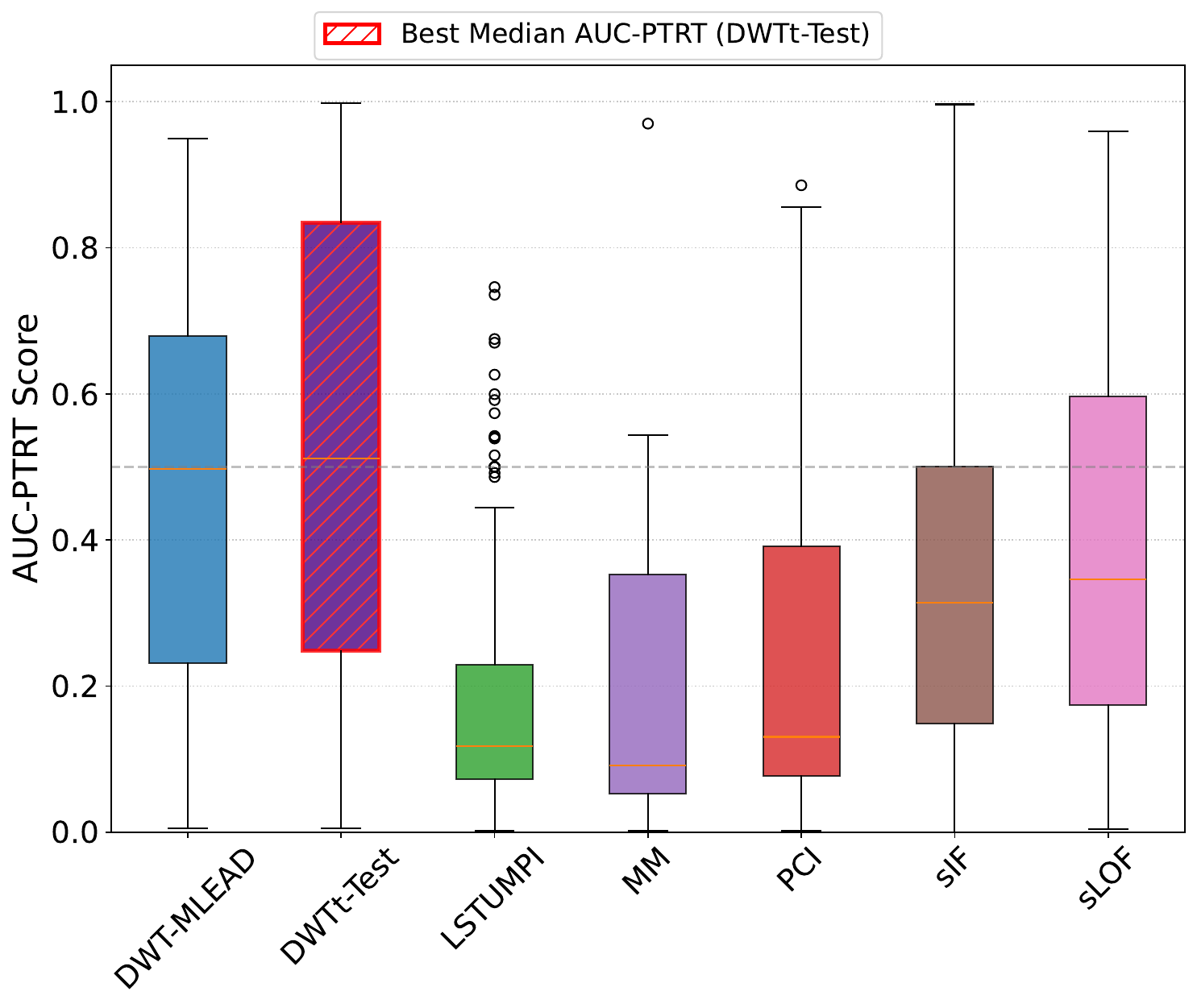}
         \caption{Distribution of AUC-PTRT scores.}
         \label{fig:PTRT_UNS_BOX}
     \end{subfigure}
     
     \caption{Quantitative evaluation of unsupervised algorithms based on AUC-PTRT. The line plot (a) shows the metric value achieved for each dataset index, while the boxplot (b) provides a statistical summary of the score distribution across the entire collection.}
     \label{fig:PTRT_UNS}
\end{figure}
Since the anomalies may behave differently, building a general-purpose unsupervised algorithm is challenging. Because of this inherent difficulty of the problem, unsupervised algorithms are usually built in order to catch a particular type of anomaly. The test datasets contain different types of real anomalies, allowing a fair evaluation of the algorithms' behaviour. In order to analyse the scores computed by these algorithms, we choose the best three performing unsupervised algorithms: DWTt-test, DWT-MLEAD and sLOF. As reported in Figure \ref{fig:SMAP_P2_UNS}, DWTt-test is able to capture the anomaly area keeping the score high for the entire anomaly window unlike sLOF, which can only catch the start and the end of the window. DWT-MLEAD detects the starting point of the anomaly window, but its score decreases over the time-steps. 
The output scores shown in Figure \ref{fig:NAB_AMBIENT_UNS} highlight the difficulty in catching both anomalies on this dataset. In particular, our algorithm detects the first anomaly window while DWT-MLEAD performs better on the second anomaly even though it exhibits a false positive around the $1500$-th time-step. The sLOF algorithm is able to catch the second anomaly window, but computes too many false positives, thus negatively affecting the result. A similar behaviour is shown in Figure \ref{fig:MSL_C2_UNS}, in which our algorithm is able to detect the first whole anomaly window while DWT-MLEAD detects only its starting points. Furthermore, the scores computed by the sLOF algorithm highlight its inefficiency with this type of anomaly. The second window seems more complicated than the previous one, causing all algorithms to fail. A very chaotic situation is shown in Figure \ref{fig:MSL_F7_UNS}. Our algorithm is able to catch all the anomalies while maintaining the scores low across the normal points. DWT-MLEAD exhibits a good behaviour on this dataset, but it's only able to detect two anomalies with one false positive around the $4000$-th time-step. sLOF fails on this dataset, computing a lot of high scores along normal data. 

A quantitative evaluation is provided in Figures \ref{fig:ROC_UNS}, \ref{fig:PR_UNS} and \ref{fig:PTRT_UNS}. The best performing algorithms are those which are based on the DWT, reaching a ROC value higher than $0.8$. Despite the similar use of DWT in the DWTt-test and DWT-MLEAD algorithms, DWTt-test performs slightly better than DWT-MLEAD, demonstrating a better adaptability to different types of outliers and confirming that the ad-hoc $t$-test detection methodology leads to better performance (additionally, as will be shown later, DWTt-test is also faster than DWT-MLEAD by an order of magnitude). The trend of the ROC metric across the datasets shown in Figure \ref{fig:ROC_UNS_CURVE} is also visible for the PR curve in Figure \ref{fig:PR_UNS_CURVE}, while the results shown in Figure \ref{fig:PTRT_UNS_CURVE} suggest a more robust behaviour of the proposed algorithm across several types of anomalies compared to the competing algorithms. The average execution times per point for all evaluated algorithms are summarised in Table \ref{tab:execution_times}. For the unsupervised methods, experiments were conducted on a MacBook Pro equipped with an Apple M1 Pro chip (8-core CPU) and 16 GB of unified memory. Empirical results demonstrate that our proposed algorithm, DWTt-test, achieves the lowest execution time among the core DWT-based methods. Notably, it outperforms DWT-MLEAD, even though the computational complexity of both algorithms is linear in the size of the input ($O(N)$). This efficiency gain is particularly relevant for real-time applications, as DWTt-test maintains a high throughput while ensuring robust detection capabilities. 

\subsection{Evaluation of Self-Supervised Algorithms}

\begin{figure}
     \centering
     \begin{subfigure}{\columnwidth}
         \centering
         \includegraphics[width=0.90\textwidth]{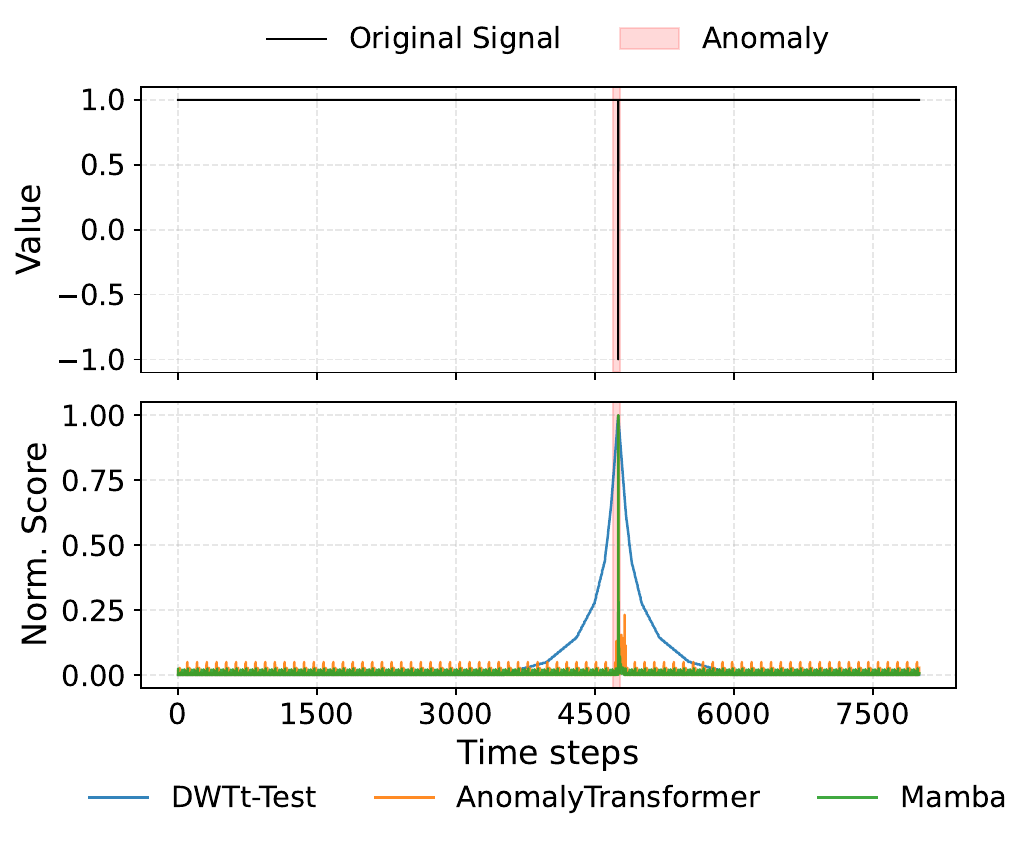}
         \caption{Anomaly scores for the A-1 dataset (SMAP collection)}
         \label{fig:SMAP_A1_SUP}
     \end{subfigure}

     \vspace{0.3cm}

     \begin{subfigure}{\columnwidth}
         \centering
         \includegraphics[width=0.90\textwidth]{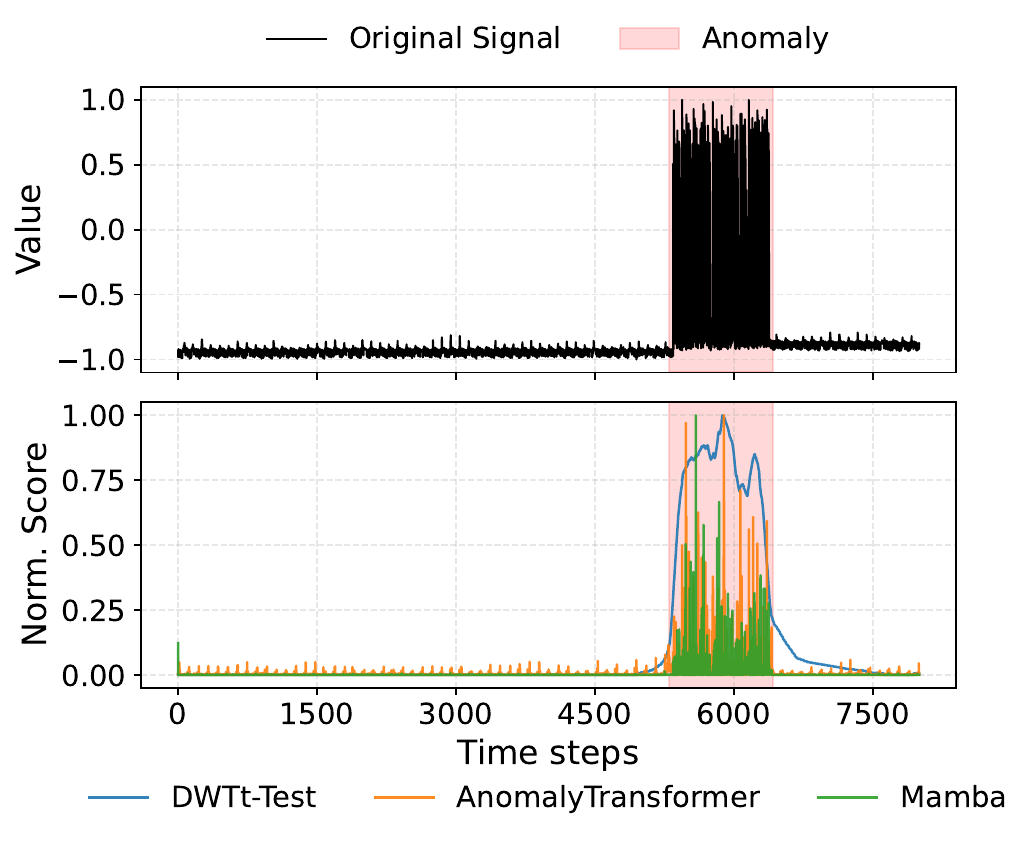}
         \caption{Anomaly scores for the P-2 dataset (SMAP collection)}
         \label{fig:SMAP_P2_SUP}
     \end{subfigure}
     
     \caption{Visual comparison of detection scores where the unsupervised DWTt-test is evaluated alongside self-supervised deep learning models, specifically AnomalyTransformer and Mamba. The subplots illustrate how each algorithm tracks anomalies across different time series from the NASA-SMAP collection.}
     \label{fig:combined_plots_sup}
\end{figure}
\begin{figure}
     \centering
     \ContinuedFloat
     
     \begin{subfigure}{\columnwidth} 
         \centering
         \includegraphics[width=0.90\textwidth]{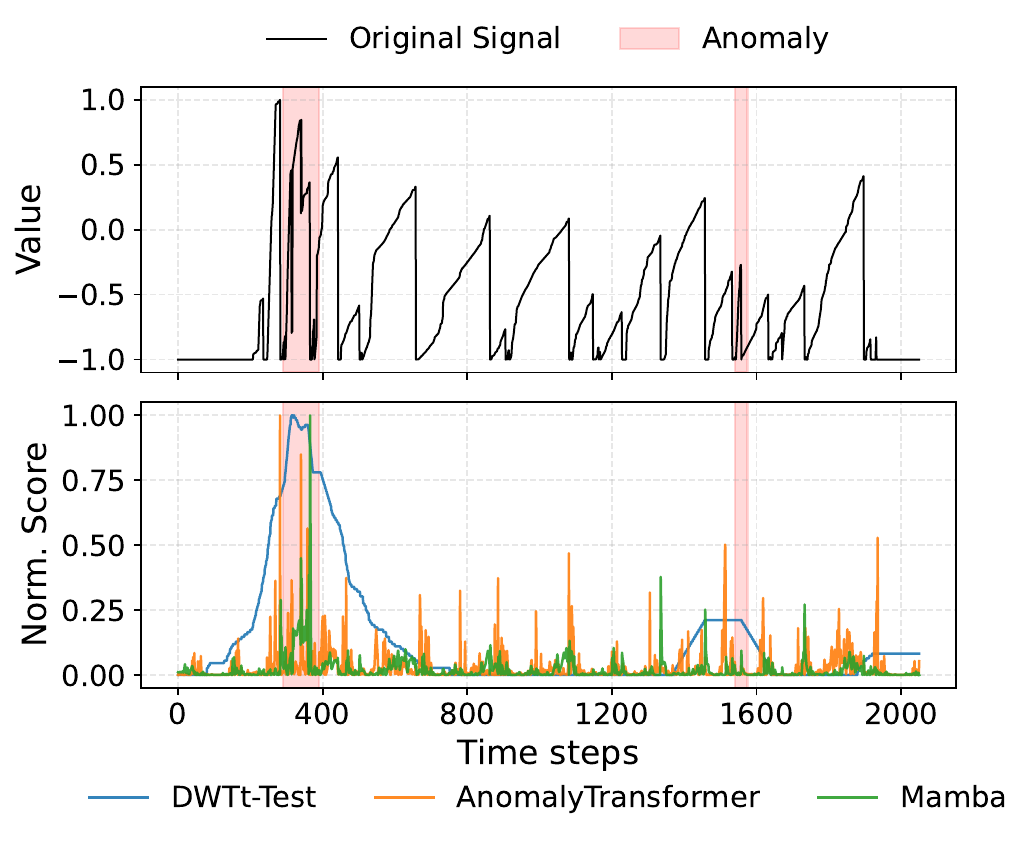}
         \caption{Anomaly scores for the C-2 dataset (MSL collection)}
         \label{fig:MSL_C2_SUP}
     \end{subfigure}

     \vspace{0.3cm}

     \begin{subfigure}{\columnwidth}
         \centering
         \includegraphics[width=0.90\textwidth]{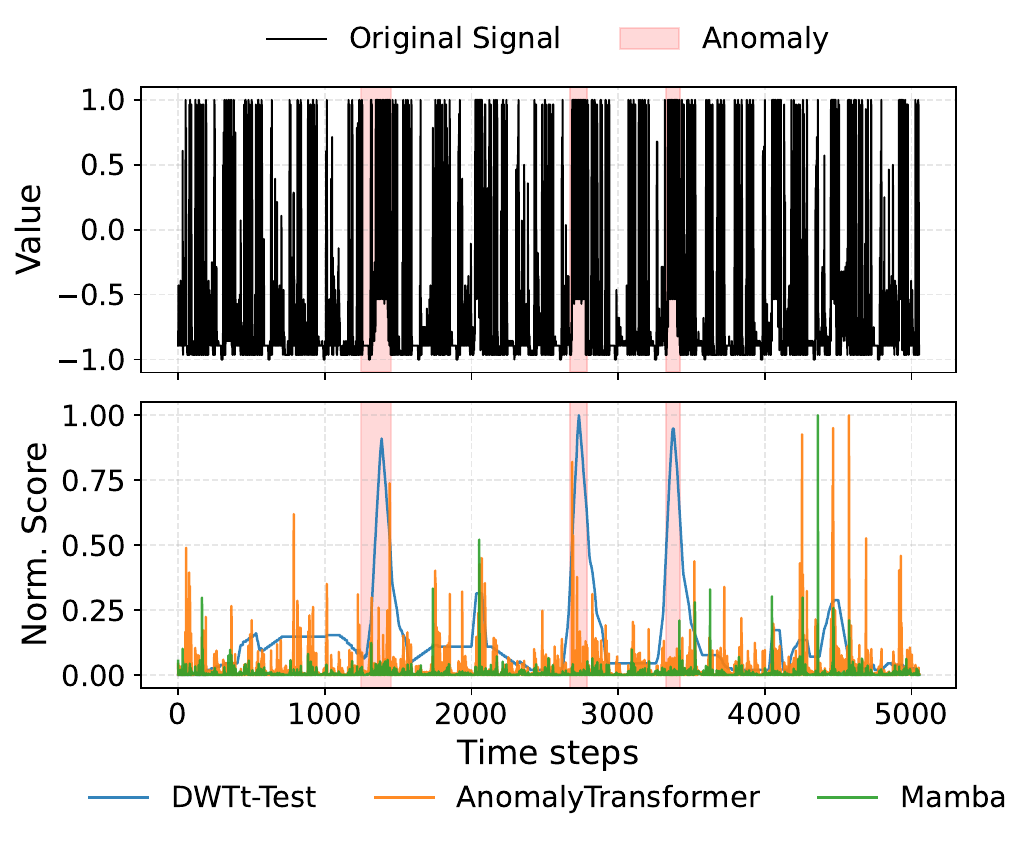}
         \caption{Anomaly scores for the F-7 dataset (MSL collection)}
         \label{fig:MSL_F7_SUP}
     \end{subfigure}
     
     \caption{Visual comparison of detection scores where the unsupervised DWTt-test is evaluated alongside self-supervised deep learning models, specifically AnomalyTransformer and Mamba. The subplots illustrate how each algorithm tracks anomalies across different time series from the NASA-MSL collection.}
\end{figure}

\begin{figure}
     \centering
     \begin{subfigure}{\columnwidth}
         \centering
         \includegraphics[width=0.95\textwidth]{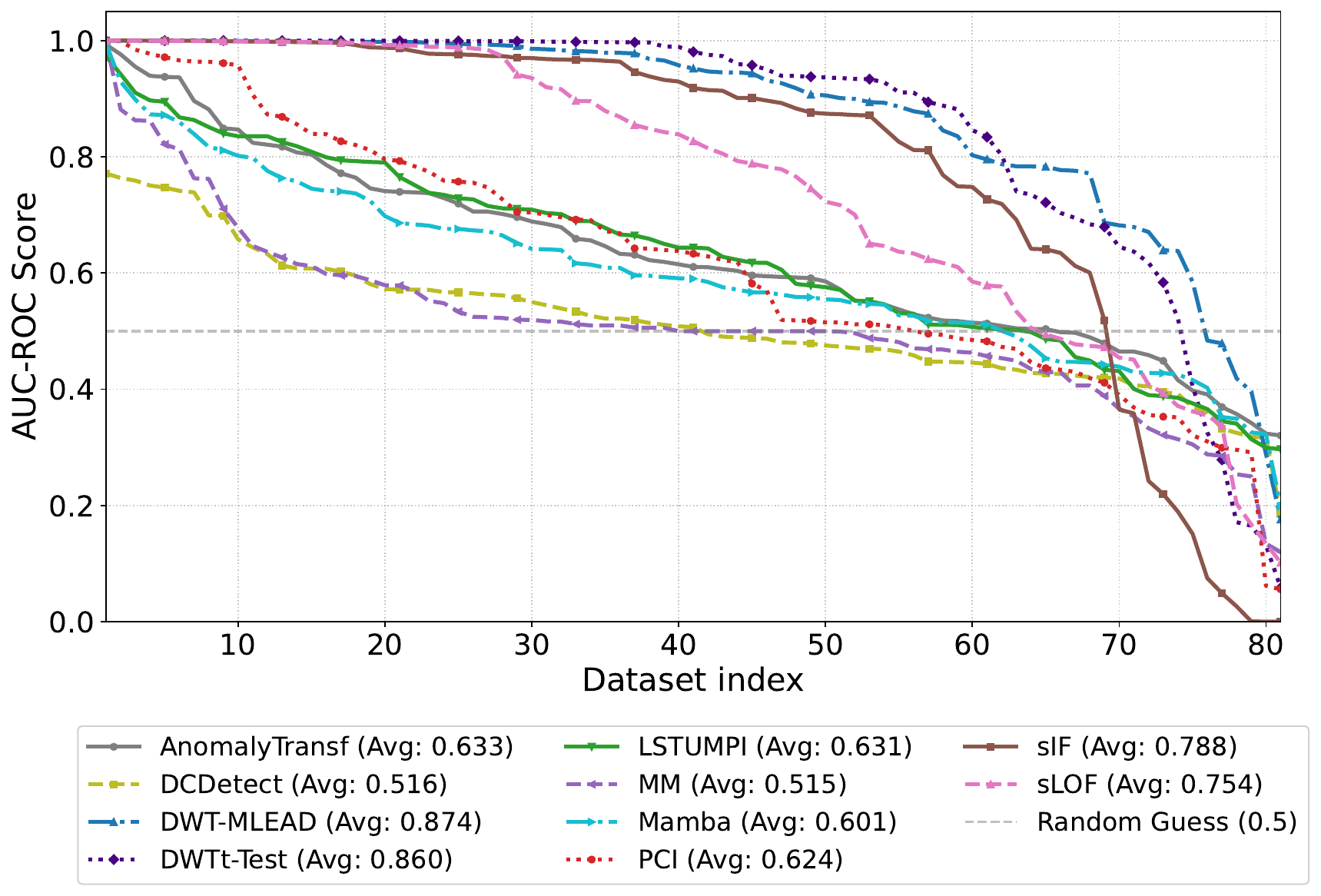}
         \caption{AUC-ROC values across NASA SMAP and MSL collections.}
         \label{fig:ROC_SUP_CURVE}
     \end{subfigure}
     
     \begin{subfigure}{\columnwidth}
         \centering
         \includegraphics[width=0.95\textwidth]{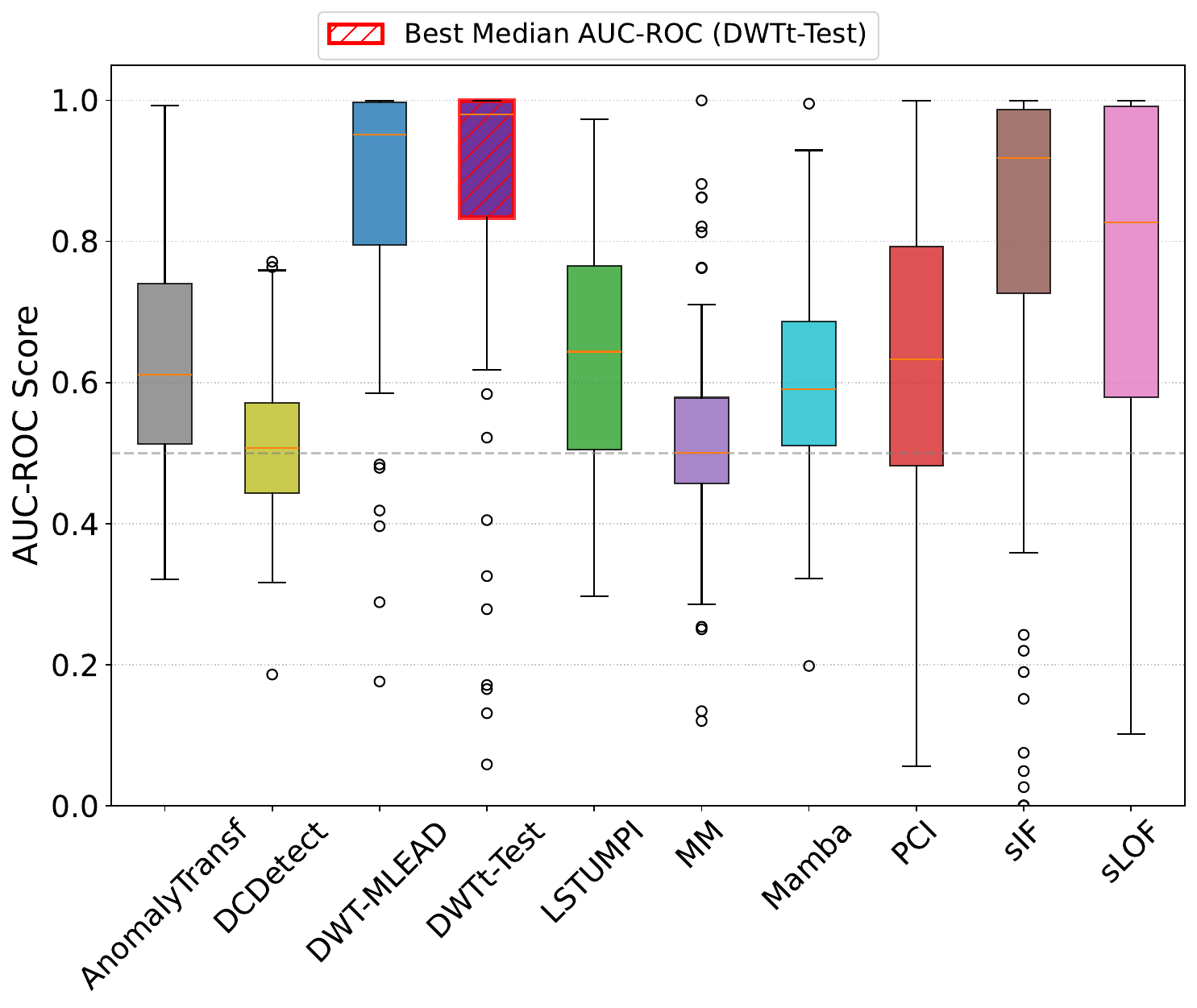}
         \caption{Distribution of AUC-ROC scores (NASA collections).}
         \label{fig:ROC_SUP_BOX}
     \end{subfigure}
     
     \caption{Quantitative evaluation of unsupervised and supervised algorithms based on AUC-ROC. The line plot (a) shows the metric value achieved for each dataset index within the NASA SMAP and MSL collections, while the boxplot (b) provides a statistical summary of the score distribution for these specific sequences.}
     \label{fig:ROC_SUP}
\end{figure}

\begin{figure}
     \centering
     \begin{subfigure}{\columnwidth}
         \centering
         \includegraphics[width=0.95\textwidth]{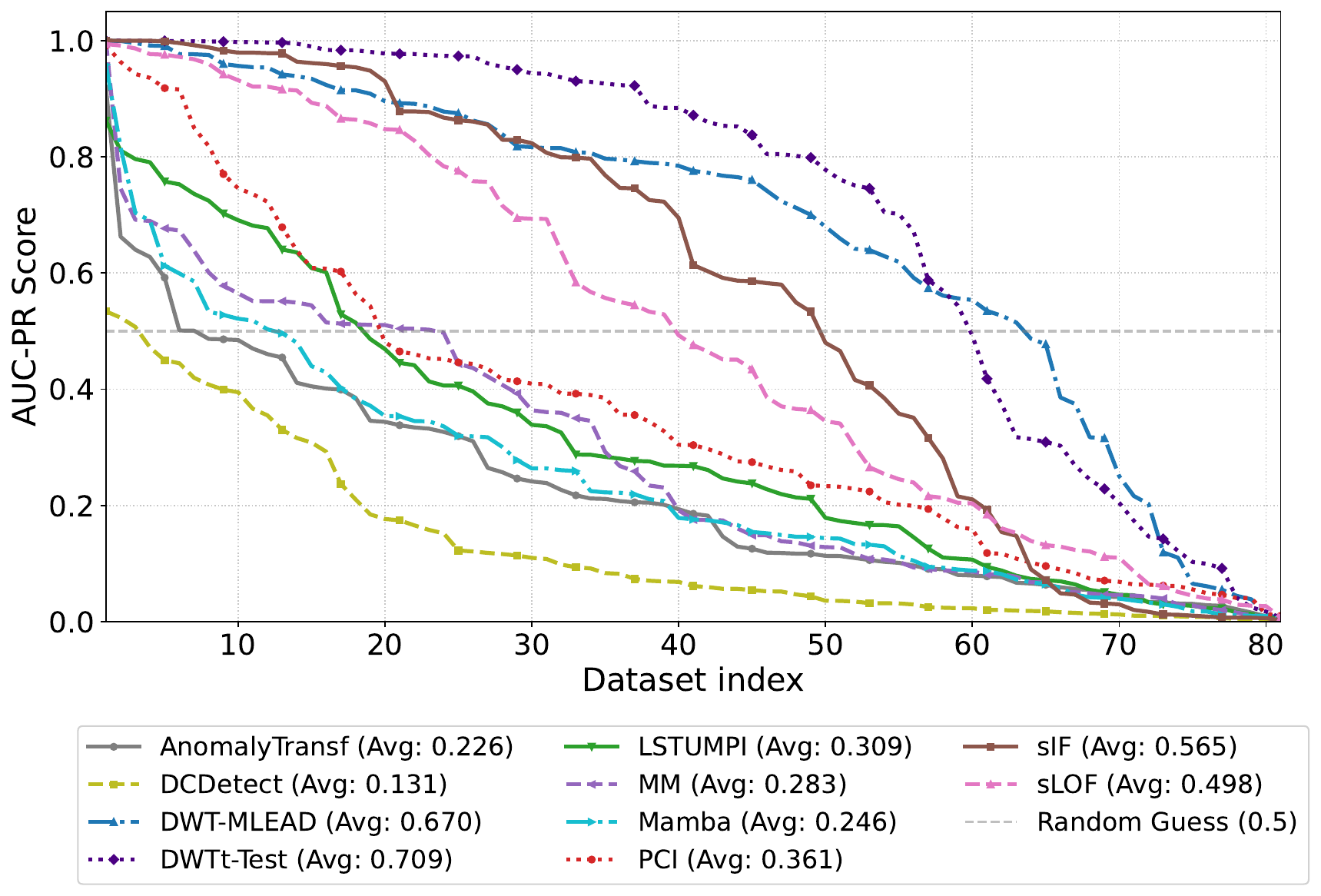}
         \caption{AUC-PR values across NASA SMAP and MSL collections.}
         \label{fig:PR_SUP_CURVE}
     \end{subfigure}
     
     \begin{subfigure}{\columnwidth}
         \centering
         \includegraphics[width=0.95\textwidth]{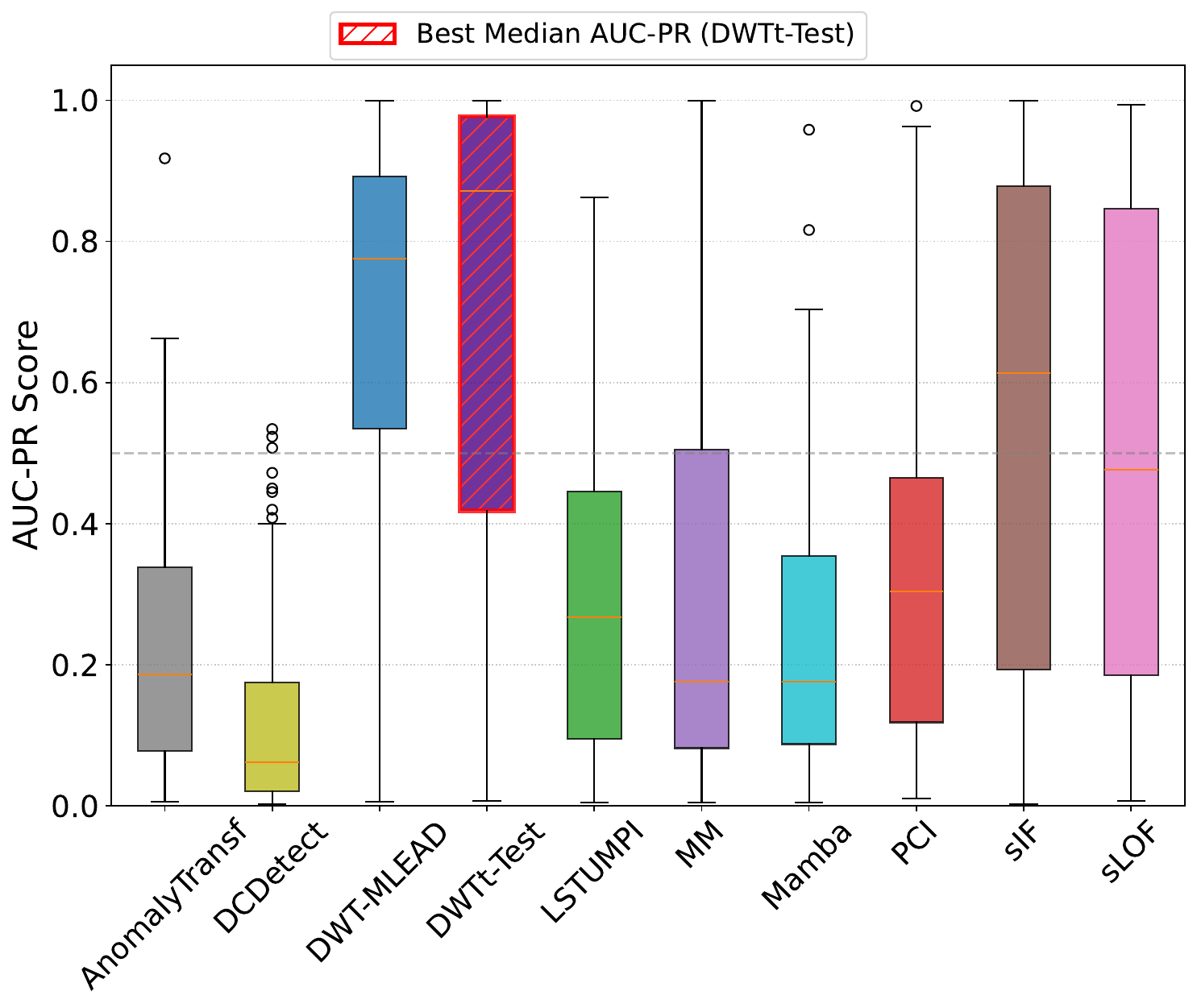}
         \caption{Distribution of AUC-PR scores (NASA collections).}
         \label{fig:PR_SUP_BOX}
     \end{subfigure}
     
     \caption{Quantitative evaluation of unsupervised and supervised algorithms based on AUC-PR. The line plot (a) shows the metric value achieved for each dataset index within the NASA SMAP and MSL collections, while the boxplot (b) provides a statistical summary of the score distribution for these specific sequences.}
     \label{fig:PR_SUP}
\end{figure}

\begin{figure}
     \centering
     \begin{subfigure}{\columnwidth}
         \centering
         \includegraphics[width=0.95\textwidth]{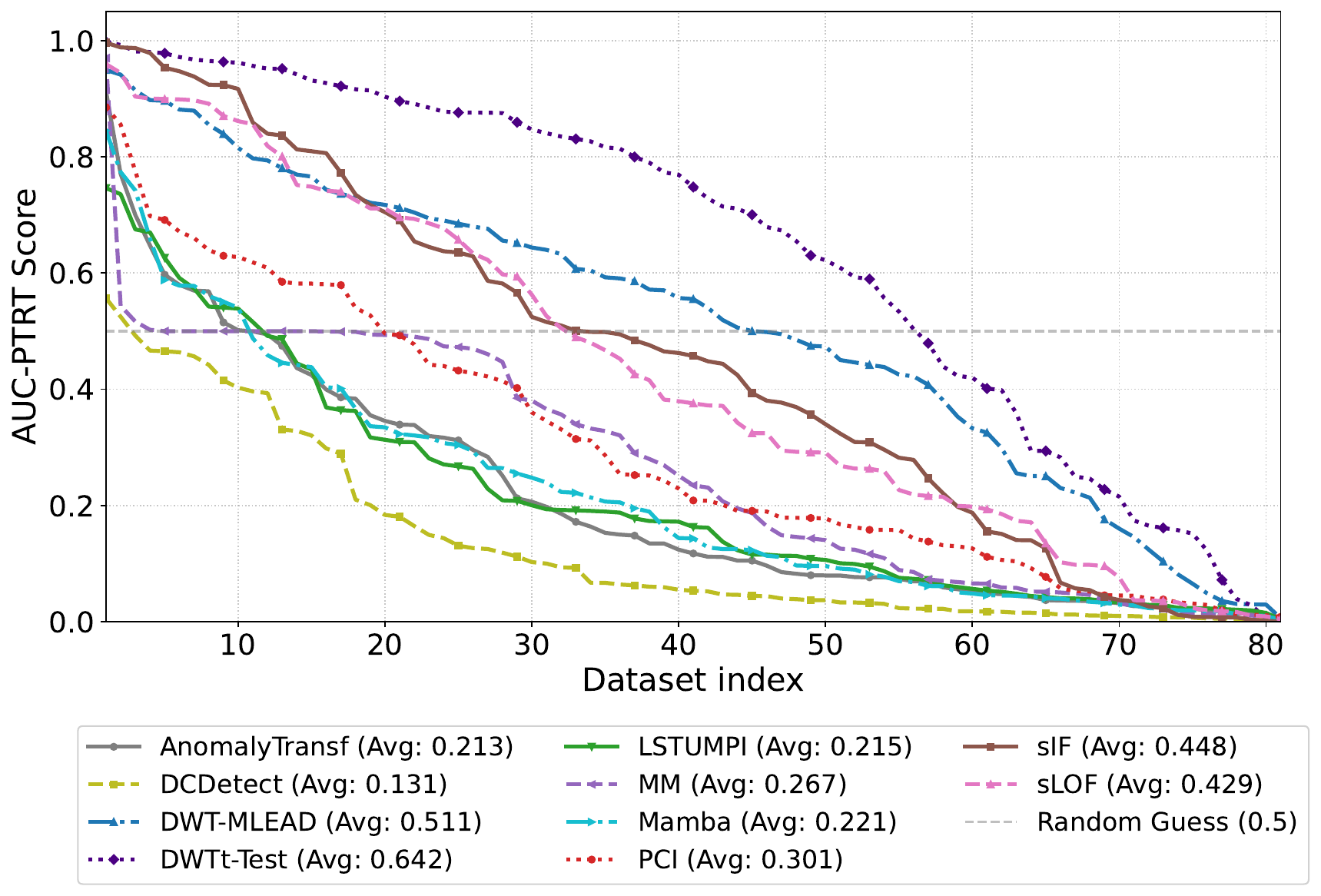}
         \caption{AUC-PTRT values across NASA SMAP and MSL collections.}
         \label{fig:PTRT_SUP_CURVE}
     \end{subfigure}
     
     \vspace{0.3cm}
     
     \begin{subfigure}{\columnwidth}
         \centering
         \includegraphics[width=0.95\textwidth]{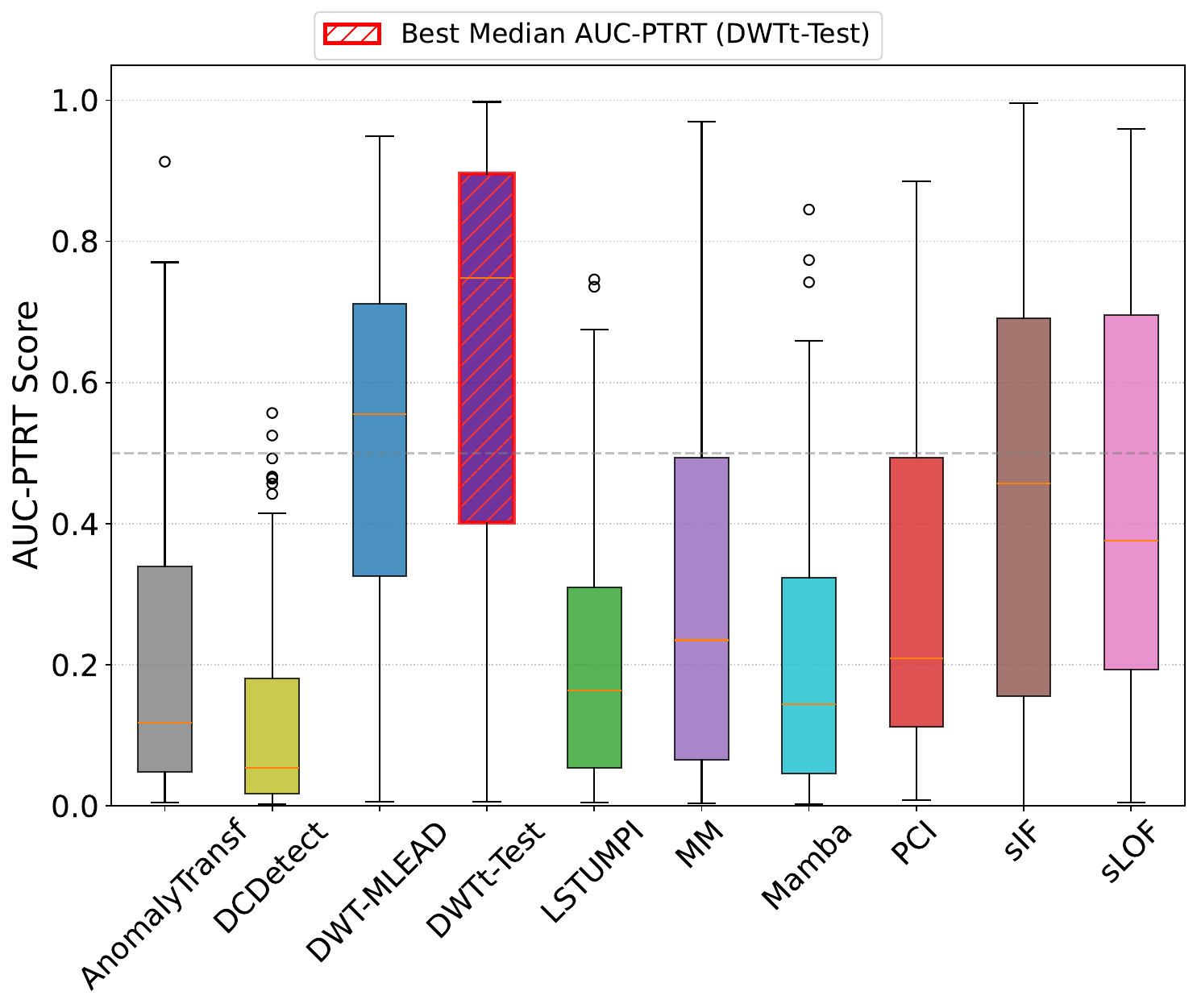}
         \caption{Distribution of AUC-PTRT scores (NASA collections).}
         \label{fig:PTRT_SUP_BOX}
     \end{subfigure}
     
     \caption{Quantitative evaluation of unsupervised and supervised algorithms based on AUC-PTRT. The line plot (a) shows the metric value achieved for each dataset index within the NASA SMAP and MSL collections, while the boxplot (b) provides a statistical summary of the score distribution for these specific sequences.}
     \label{fig:PTRT_SUP}
\end{figure}

After training the self-supervised models on the NASA-SMAP and NASA-MSL, these deep learning models were compared against the unsupervised counterparts on the corresponding test datasets. Although the models were trained following the methodology developed by the authors, these models produced poor performances compared to the unsupervised algorithms. In order to better understand these results, an analysis of the output scores is needed. Since the two best self-supervised models are Mamba and AnomalyTransformer, we reported only their scores to have a better, uncluttered visualisation of the model behaviours. Figures \ref{fig:SMAP_A1_SUP} and \ref{fig:SMAP_P2_SUP} represent baseline datasets useful for evaluating the proper functioning of the deep learning models. In particular, Figure \ref{fig:SMAP_A1_SUP} illustrates how these models operate correctly in the presence of a spike anomaly. Although all the models under test detect the anomaly (yielding the same performance), the deep learning models are more precise, increasing the outlier scores only at the exact anomaly time step, unlike the wider area covered by the proposed approach. Further evidence of these “irregular” score behaviours is shown in Figure \ref{fig:SMAP_P2_SUP}. All the computed scores inside the anomaly window are generally higher than the normal scores, underlining  that the deep learning models are able to catch simple outliers like these. The final AUC score computed on this dataset is lower than the AUC score of the DWTt-test, because of the consistency of the scores. Indeed, the output scores generated by the models are highly irregular, yielding more false negatives. A more complex dataset is shown in Figure \ref{fig:MSL_C2_SUP}. The first anomaly is detected by all the models, meanwhile the second one is more challenging to catch. The main differences are depicted across the normal points. In particular, the AnomalyTransformer model shows a more unstable behaviour compared to the DWTt-test and the Mamba model, leading to many false positives. The irregularities within the dataset shown in Figure \ref{fig:MSL_F7_SUP} test the structural limitations of the Mamba model, complicating the extraction of meaningful patterns, leading to zero outliers detected against the two outliers detected by the AnomalyTransformer model. Neither model is able to maintain high scores for all outlier windows, unlike the proposed algorithm that performs better despite the longer outlier sequences. Furthermore, the deep learning models generate false positives between the $4000$-th and $5000$-th time-steps. Overall, the deep learning models are able to identify anomaly sequences, but because of the significant irregularities of the outlier scores, the results are poor, leading to a large number of false positives or false negatives, negatively affecting the overall performance. 

The quantitative summary results are available in Figures \ref{fig:ROC_SUP}, \ref{fig:PR_SUP} and \ref{fig:PTRT_SUP}. In particular, the latter analysis is evident from Figures \ref{fig:PR_SUP_CURVE} and \ref{fig:PTRT_SUP_CURVE}, showing that the deep learning models are not able to maintain consistency across the anomaly windows, requiring the PA (Point Adjustment) protocol to boost performance. The execution times for the self-supervised models, specifically Mamba, AnomalyTransformer, and DC-Detector, are reported in Table \ref{tab:execution_times}. These models were evaluated using an NVIDIA RTX 4090 GPU to leverage their deep learning architectures. It is important to emphasise that these GPU-based results are not directly comparable to the execution times of the unsupervised algorithms, which were measured on a CPU-based architecture (Apple M1 Pro). While the parallel processing capabilities of the RTX 4090 GPU allow these complex models to achieve competitive per-point latency, the underlying hardware disparity must be considered when evaluating overall efficiency. However, even with the advantage of high-end GPU acceleration, our unsupervised DWTt-test remains highly competitive, demonstrating its suitability for deployment in environments where dedicated graphics hardware may not be available.
\begin{table}
    \centering
    \caption{Comparison of execution times per point (seconds). Unsupervised algorithms are evaluated on an Apple M1 Pro CPU, while self-supervised models are evaluated using a NVIDIA RTX 4090 GPU.}
    \label{tab:execution_times}
    \begin{tabular}{l l S[table-format=1.2e-1]}
        \toprule
        \textbf{Category} & \textbf{Algorithm} & {\textbf{Time per Point (s)}} \\
        \midrule
        \textit{Unsupervised} & MedianMethod & 6.00e-8 \\ 
        (CPU) & \textbf{DWTt-test} & 3.50e-6 \\
        & PCI & 5.50e-6 \\
        & sIF & 1.50e-5 \\
        & sLOF & 1.80e-5 \\
        & DWT-MLEAD & 5.00e-5 \\
        & LSTUMPI & 5.00e-4 \\
        \midrule
        \textit{self-supervised} & DC-Detector & 1.22e-5 \\
        (GPU) & AnomalyTransformer & 1.42e-5 \\
        & Mamba & 1.62e-5 \\
        \bottomrule
    \end{tabular}
\end{table}

\section{Conclusions}
\label{conclusions}

In this paper, we introduced DWTt-test, a novel and highly efficient unsupervised algorithm for anomaly detection in univariate time series. By synergistically combining the Haar Discrete Wavelet Transform (DWT) with a rigorously derived, ad-hoc statistical $t$-test, our approach accurately isolates anomalous patterns across multiple resolution levels without the need for labeled training data. Furthermore, we provided formal mathematical proofs demonstrating that our tailored $t$-score rigorously follows a Student's $t$-distribution under the algorithm's operational settings.

Extensive experimental evaluations were conducted across diverse, real-world benchmark datasets, for a total of 343 datasets. To ensure a fair and rigorous comparison, we strictly adhered to a threshold-agnostic and point-adjustment-free evaluation protocol, reflecting the true inherent discriminative power of the models. The empirical results demonstrate that DWTt-test significantly outperforms state-of-the-art unsupervised baselines across multiple robust evaluation metrics, including AUC-ROC, AUC-PR, and AUC-$P_T R_T$. Moreover, our comparative analysis against recent self-supervised deep learning architectures revealed critical vulnerabilities in these state-of-the-art approaches. When evaluated without the overly optimistic point-adjustment protocol, these deep learning models often produced highly irregular anomaly scores, resulting in a substantial number of false positives and false negatives. Conversely, DWTt-test maintained consistent and reliable scoring across all the anomalous windows.

From a computational perspective, the proposed algorithm exhibits a strict linear time complexity, $\mathcal{O}(N)$. It achieved the lowest execution time per point among the unsupervised methods on standard CPU hardware (excluding MedianMethod, which is highly inaccurate; notably, DWTt-test is also faster than DWT-MLEAD by an order of magnitude), bypassing the need for expensive GPU acceleration required by deep learning competing models. This exceptional efficiency, coupled with high detection accuracy, makes DWTt-test exceptionally well-suited for real-time monitoring and deployment in resource-constrained IoT or edge computing environments.

Future work will focus on parallelising and extending the DWTt-test mathematical framework and hierarchical tree structure to directly process multivariate time series, further expanding its applicability to complex, multi-sensor industrial environments.

\section*{\textbf{Declaration of competing interest}} The authors declare that they have no known competing financial interests or personal relationships that could have appeared to influence the work reported in this paper.

\section*{\textbf{Data availability}}
All the datasets used in this work have been preprocessed and made available by \cite{Schmidl2022}.

\section*{\textbf{Code availability}}
The code used in this work will be publicly released in case of acceptance of the manuscript.

\printcredits


\bibliographystyle{cas-model2-names}
\bibliography{bibliography}

\end{document}